\documentclass[accepted]{uai2022} % after acceptance, for a revised
\usepackage[american]{babel}
\usepackage{natbib} % has a nice set of citation styles and commands
\usepackage{mathtools} % amsmath with fixes and additions
\usepackage{booktabs} % commands to create good-looking tables
\usepackage{tikz} % nice language for creating drawings and diagrams
\usepackage[utf8]{inputenc} % allow utf-8 input
\usepackage{booktabs}       % professional-quality tables
\usepackage{amsfonts,amsmath,amssymb,amsthm}       % blackboard math symbols
\usepackage{nicefrac}       % compact symbols for 1/2, etc.
\usepackage{microtype}      % microtypography
\usepackage{xcolor}         % colors
\usepackage{dsfont}
\usepackage{bm}
\usepackage{siunitx}
\usepackage[ruled]{algorithm2e}
\DeclareUrlCommand\UScore{\urlstyle{rm}}
\newtheorem{theorem}{Theorem}
\newtheorem{definition}{Definition}
\usepackage[capitalise]{cleveref}

\newcommand{\ie}{i.e.~}

\newcommand{\beq}[1]{\begin{equation} \eqlab{#1}}
\newcommand{\eeq}{\end{equation}}
\newcommand{\bal}{\begin{align}}
\newcommand{\eal}{\end{align}}

\newcommand{\bsub}{\begin{subequations}}
\newcommand{\esub}{\end{subequations}}

\newcommand{\eqlab}[1]{\label{eq:#1}}
\renewcommand{\eqref}[1]{Eq.~(\ref{eq:#1})}

\newcommand{\figlab}[1]{\label{fig:#1}}
\newcommand{\tabref}[1]{Table~\ref{tab:#1}}

\newcommand{\tablab}[1]{\label{tab:#1}}
\newcommand{\appref}[1]{Appendix~\ref{app:#1}}

\newcommand{\grad}{\boldsymbol{\nabla}}

\renewcommand{\vec}[1]{\bm{#1}}

\newcommand{\mL}{\mathcal{L}}
\newcommand{\mO}{\mathcal{O}}

\newcommand{\mA}{\mathcal{A}}

\newcommand{\mU}{\mathcal{U}}

\newcommand{\real}{\mathbb{R}}

\newcommand{\br}[1]{\left\lbrack #1 \right\rbrack}
\newcommand{\paren}[1]{\left(#1\right)}
\newcommand{\tub}[1]{\left\{#1\right\}}
\newcommand{\mr}[1]{\mathrm{#1}}
\newcommand{\vt}[1]{\left.#1\right\vert}
\newcommand{\abs}[1]{\left\lvert#1\right\rvert}

\newcommand{\aaa}{\vec{a}}

\newcommand{\vvv}{\vec{v}}

\newcommand{\xxx}{\vec{x}}

\newcommand\xlat{\xxx_{\mr{lat}}}
\newcommand\xobs{\xxx_{\mr{obs}}}

\newcommand{\meanp}[2]{\mathbb{E}_{#1}\br{#2}}
\newcommand{\kl}[2]{\mr{KL}\paren{\vt{\vt{#1~}}#2}}

\hypersetup{colorlinks=false, allcolors=black, pdfborder={0 0 0}}
\title{Probabilistic Surrogate Networks for Simulators with Unbounded Randomness}

\author[1]{\href{mailto:<amunk@cs.ubc.ca>?Subject=Probabilistic Programming, Surrogate Modeling}{Andreas Munk}}
\author[1]{Berend Zwartsenberg}
\author[1]{Adam \'Scibior$^{2,}$}
\author[3]{At{\i}l{\i}m G{\"u}ne{\c s} Baydin}
\author[4]{Andrew Stewart}
\author[4]{\qquad Goran Fernlund}
\author[5]{Anoush Poursartip$^{4,}$}
\author[1]{Frank Wood$^{2,6,}$}
\affil[1]{%
  Department of Computer Science\\
  University of British Columbia
  \quad
  $^{2}$Inverted AI Ltd.
}

\affil[3]{%
  Department of Engineering Science\\
  University of Oxford
}
\affil[4]{%
    Convergent Manufacturing Technologies Inc.
}
\affil[5]{%
   Composites Research Network\\
   University of British Columbia
   \quad
   $^{6}$Mila, CIFAR AI Chair
}

\begin{document}
\maketitle

\begin{abstract}
  We present a framework for automatically structuring and training fast,
  approximate, deep neural surrogates of stochastic simulators. Unlike
  traditional approaches to surrogate modeling, our surrogates retain the
  interpretable structure and control flow of the reference simulator. Our
  surrogates target stochastic simulators where the number of random variables
  itself can be stochastic and potentially unbounded. Our framework further
  enables an automatic replacement of the reference simulator with the surrogate
  when undertaking amortized inference. The fidelity and speed of our surrogates
  allow for both faster stochastic simulation and accurate and substantially
  faster posterior inference. Using an illustrative yet non-trivial example we
  show our surrogates' ability to accurately model a probabilistic program with
  an unbounded number of random variables. We then proceed with an example that
  shows our surrogates are able to accurately model a complex structure like an
  unbounded stack in a program synthesis example. We further demonstrate how our
  surrogate modeling technique makes amortized inference in complex black-box
  simulators an order of magnitude faster. Specifically, we do simulator-based
  materials quality testing, inferring safety-critical latent internal
  temperature profiles of composite materials undergoing curing.
\end{abstract}

\section{Introduction}
\label{sec:introduction}

Stochastic simulators are accurate generative models that encode the
relationship between random variables. Simulators can be used to reason about
the relationship between latent variables and real world observations which the
simulator is assumed to accurately model. Whether in aeronautical
engineering~\citep{wu2018}, nonlinear flow physics~\citep{veldman2007a},
finance~\citep{raberto2001a}, or modeling the brain's blood
flow~\citep{perdikaris2016a}, simulators play an important role in design,
diagnosis, and manufacturing. Unfortunately, complex simulators are often
computationally expensive, ruling them out for just-in-time uses. This problem
is exacerbated in stochastic simulators as these often need to be run many times
to accurately estimate quantities of interest. A natural solution to this
problem, known as surrogate modeling, is to construct a fast approximation to
the reference simulator. This approach has found success in applications in
various fields including computational fluid dynamics (CFD)~\citep{glaz2010a},
aerospace engineering~\citep{jeong2005efficient}, material
science~\citep{rikards2004surrogate} and quantum chemistry~\citep{gilmer2017a}.
These surrogates learn to approximate the input to output mapping represented by
a simulator, usually by fitting a regressor to samples drawn from the simulator.
However, when such a simulator is stochastic, and especially when the number of
internal random variables is unbounded, it is not immediately clear how to
extend these ideas. It becomes impossible to write down a predefined
parametrization. This is exactly the issue our work addresses: to provide a
framework for surrogate modeling in the case where the number of random
variables is generally unbounded.

Stochastic simulators can come in the form of (I) deterministic simulators with
a fixed-dimensional vector of randomly distributed inputs (equivalent to a
push-forward or structural equation model) or (II) a program that uses random
variables internally. Constructing a surrogate for simulators that consumes
randomness internally (type II) is done in one of two ways: (1) All internally
utilized random variables are externalized and specified a priori as inputs,
effectively transforming a type II stochastic simulator into type I. (2)
Internal random variables are implicitly marginalized over, in which case any
structural information internal to the simulators will be lost. In particular,
considering (1), the randomness of a stochastic simulator can be abstracted to a
single random number seed. Alternatively, and less extreme, samples of all
random variable types can be obtained by deterministically transforming
$\mU(0,1)$ pseudorandom numbers. So we can in theory transform a stochastic
simulator of type II into a stochastic simulator of type I with $\mU(0,1)$
distributed inputs. However, identifying all the internal random variables is in
general impossible when a Turing complete language is used to specify the type
II stochastic simulator that uses looping, branching, and other control flow
constructs. This is because one must be able to identify or ``address'' all the
random variables in advance. This is infeasible as the space of random variables
can be countably infinite. So, any generic scheme to externalize the random
variables of a type II stochastic simulator will involve
\emph{under-approximating} the original stochastic simulator, as only a finite
number of variables can be externalized. Further like for option (2), such
externalization discards structural information about the relationship between
the otherwise internal random variables. As it is known that utilizing
information about the structure of the model enables it to generalize better and
lead to lower estimation loss~\citep{bishop2006a}, it is desirable to retain all
such information.

To address this we introduce a novel approach to surrogate modeling which
captures and fully utilizes the structure of the stochastic simulator. Our
surrogates, which we call \textit{probabilistic surrogate networks} (PSNs), do
not suffer from being an under-approximation. However, as we will discuss, it
might be desirable to choose to execute them as under-approximations for
practical purposes. Particularly, by framing the surrogate modeling problem in
the context of probabilistic programming, our model architecture automatically
replicates distributions over traces for a given reference simulator. This enables
PSNs to generate interpretable sequences of latent variables that are
fully compatible with the reference simulator. We achieve this by simultaneously
learning to approximate the latent probability distributions \textit{and} the
control flow of the original simulator. Our method therefore targets simulators
of type II in addition to type I, and we emphasize that it can handle simulators
with arbitrarily many random variables. As a corollary we introduce a novel
method for parameterizing a classifier defined over an unbounded number of
classes.

Faster simulation via surrogate modeling is in itself useful. However, the
speedup PSNs provide arguably has even greater impact on the
``inversion'' of simulators. Here inverting a simulator means performing
Bayesian inference over latent variables given observed values of outputs. This
definition blurs the line between stochastic simulators and probabilistic models
and should be considered a key point of probabilistic
programming~\citep{baydin2019etalumis}. In this paper we illustrate how PSNs
leverage faster inference by employing them in conjunction with the neural
network based \textit{inference compilation} (IC)
framework~\citep{le2017inference} and its PyProb~\citep{pyprob} realization.
PyProb is a probabilistic programming language (PPL) that enables Bayesian
inference in stochastic simulators written in other programming
languages~\citep{ppx}, by intercepting and controlling random number draws
during simulator execution. This process is explained elsewhere in full
technical detail~\citep[chapt.~6]{van2018introduction}. PyProb was chosen due to
several desirable features, such as automatic address construction.

\section{Background}

\subsection{Probabilistic Programming}

The probabilistic programming paradigm equates a generative model with a program
written in a probabilistic programming language (PPL). An inference backend
takes the program and observed data and generates inference results, usually in
the form of samples from a posterior distribution. PPLs can be broadly
categorized as restricted, which limit the set of expressible models to ensure
that particular inference algorithms can be
applied~\citep{lunn2009bugs,InferNET18,milch2005a,carpenter2017stan,tran2016edward},
and unrestricted (universal), which allow arbitrary
models~\citep{goodman2008a,mansinghka2014a,wood2014new,pfeffer2009figaro,dippl,bingham2018pyro}.
For instance, universal PPLs allow programs to contain for-loops where the
number of iterations itself is stochastic and unbounded. For our purposes it is
particularly important to note that extending an existing Turing-complete
programming language with operations for sampling and conditioning results in a
universal PPL~\citep{dippl}. For this reason existing stochastic
simulators written in Turing-complete languages are programs in a universal PPL.
As PSNs target universal PPLs we focus our discussion here on those.

A crucial concept is that of a \emph{trace} of a probabilistic program. A trace
is a sequence of random variables $(x_{a_t}, a_t)$ for $t=1,\dots,T$, where
$a_t\in\mA$ is an address~\citep{wingate2011a} of a random variable and
$x_{a_t}$ its value. $\mA = \{\alpha_1, \alpha_2, \dots\}$ is a countable
(potentially infinite) set of possible address values, which uniquely identify
all random variables the simulator could ever produce. The purpose of the
addresses is to identify the same random variables across different execution
traces to facilitate correct inference. The trace length $T$ can vary between
different executions of the same program and is generally unbounded.

Every probabilistic program specifies a joint distribution over the space of
traces. Defining $\xxx=(x_{a_1}, \dots, x_{a_T})$ and $\aaa=(a_1,\dots,a_T)$
this distribution is denoted
\begin{equation} \label{eq:prior}
  p(\xxx,\aaa) = \prod_{t=1}^T p(a_t|x_{<a_{t}},a_{<t})
  p(x_{a_t}|x_{<a_{t}},a_{\leq t}),
\end{equation}
where $x_{<a_t}=\tub{x_{a_{t'}}|x_{a_{t'}}\in\xxx,t'<t}$, $a_{<t}=a_{0:t-1}$,
$a_{\leq t}=a_{0:t}$, $a_{0}$ being the \texttt{begin-execution} address and
$x_{<a_{1}}=\emptyset$. For each $t$, $p(a_t|x_{<a_{t}},a_{<t})$ is the address
transition probability distribution and $p(x_{a_t}|x_{<a_{t}},a_{<t})$ is the
distribution passed to the \texttt{sample} or \texttt{observe} statements in the
program. It is these statements which allow for automatic inference in PPLs. The
subset of $\xxx$ specified by \texttt{observe} statements is denoted $\xobs$,
while the remaining variables are denoted $\xlat$
-- i.e. those specified using \texttt{sample} statements. The goal of inference
is to compute the posterior distribution
$p(\xlat|\xobs)=\sum_{\aaa} p(\xlat,\aaa|\xobs)$. It should be noted that in
probabilistic programs $\aaa$ is always deterministic when conditioned on $\xxx$
making the marginalization of $\aaa$ trivial, as
$p(a_{t}|x_{<a_{t}},a_{<t})=\delta{\paren{a_{t}-f(x_{<a_{t}},a_{<t})}}$, where
$\delta\paren{\cdot}$ is the Kronecker delta function and $f(\cdot)$ is the deterministic
function defined by the simulator which specifies the address transition from
address $a_{t}$ given $x_{<a_{t}},a_{<t}$. However, modeling $\aaa$ as a random
variable is essential to our PSN construction.

\subsection{Inference Compilation}
\label{sec:ic}

Inference compilation (IC)~\citep{le2017inference} is an amortized algorithm for
performing inference in probabilistic programs using sequential importance
sampling (SIS). It works by constructing an \emph{inference network}, which
constructs proposal distributions for all the latent random variables in the
program, conditioned on the observed variables.

IC is essentially a self-normalizing importance sampler, specifically developed
as an inference engine for probablistic programming languages. IC infers
$p(\xlat|\xobs)$ using a proposal distribution $q(\xlat|\xobs)$. It draws $K$
samples $\xlat^k \overset{i.i.d.}{\sim}q$, computes the weights
$w^k = \frac{p(\xlat^k,\xobs)}{q(\xlat^k|\xobs)}$, and approximates
$p(\xxx_{\mr{lat}}|\xxx_{\mr{obs}})\approx\nicefrac{\sum_{k=1}^{K}w^k\delta(\xxx_{\mr{lat}}^k-\xxx_{\mr{lat}})}{\sum_{k=1}^{K}w^k}$.

The proposal distribution $q$ factorizes in $t$ just like $p$. Subsequent
conditional distributions in $q$ are constructed using a recurrent deep neural
network, called the inference network. Specifically,
\begin{align}
  q_\phi(\xxx_{\mr{lat}},\aaa|\xxx_{\mr{obs}}) = \prod_{\scriptscriptstyle x_{a_t}^{\mr{lat}}\in\xxx_{\mr{lat}}}
  &q(x^{\mr{lat}}_{a_t}|\eta_{a_t}(x^{\mr{lat}}_{<a_t},a_{<t},\xxx_{\mr{obs}},\phi))\nonumber\\
  &\times \prod_{t=1}^{T}q(a_{t}|x_{<a_{t}},a_{<t}),\eqlab{ic-proposal}
\end{align}
where $x^{\mr{lat}}_{<a_t}=\tub{x_{a_{t'}}|x_{a_{t'}}\in\xxx_{\mr{lat}},t'<t}$,
$\phi$ are the parameters of the inference network, and $\eta_{a_t}(\cdot)$ is
the function computed by the neural network. We emphasize here how we explicitly
write the address transitions as part of the inference problem, but note that in
IC (and other similar inference engines) the address transitions in the
posterior are defined as
$q(a_{t}|x_{<a_{t}},a_{<t})\equiv p(a_{t}|x_{<a_{t}},a_{<t})$.

The proposal $q_{\phi}$ is trained to match the true posterior
$p(\xxx_{\mr{lat}},\aaa|\xxx_{\mr{obs}})\propto p(\xxx,\aaa)$, where the
distance between the posteriors $p$ and $q_{\phi}$ is measured using the
Kullback--Leibler (KL) divergence $\kl{p}{q_{\phi}}$. In order to match
$q_{\phi}$ for all possible $\xxx_{\mr{obs}}$ the expected KL divergence under
the marginal $p(\xxx_{\mr{obs}})$ is minimized.

It should be emphasized here, that for inference engines where
$q(a_{t}|x_{<a_{t}},a_{<t})\equiv p(a_{t}|x_{<a_{t}},a_{<t})$, the program and
inference engine must run \textit{concurrently}. That is, the address
transitions are provided by the program via sampling from the dirac distribution
$p(a_{t}|x_{<a_{t}},a_{<t})=\delta\paren{a_{t}-a'_{t}}$, where
$a'_{t}=f(x_{<a_{t}},a_{<t})$ is the deterministic address given
$x_{<a_{t}},a_{<t}$. This has two implications: (1) any surrogate modeling
framework incorporated into a PPL framework must be able to provide such address
transitions. (2) The runtime of inference engines relying on executing the
reference simulator, like IC, will be computationally constrained by the
computational complexity of the reference simulator. Such cases would be
examples where surrogate models, like PSNs, in PPLs can drastically speed up the inference procedure.

\section{Probabilistic Surrogate Networks}
\label{sec:psn}

PSNs are constructed to model a distribution over the trace space. They will
replace the original program, thereby facilitating faster simulation and
inference, provided the PSN is faster than the original program. PSNs factorize
identically to the distribution of the original program specified
in~\eqref{prior}. Specifically, the distribution represented by a PSN is defined
as
\begin{align}
  s_\theta(\xxx,\aaa)=\prod_{t=1}^{T}s(x_{a_t}|\xi_{a_t}(x_{<a_{t}},a_{\leq t};\theta)) \\
  \times s(a_t| \zeta_{a_{t-1}} (x_{<a_{t}},a_{<t};\theta)), \eqlab{surr-model}
\end{align}
where $\xi_{a_t}(\cdot)$ and $\zeta_{a_{t-1}}(\cdot)$ are neural networks. At the center of our PSNs, there is a recurrent neural network
(RNN) that enables the density of $x_{a_{t}}$ to depend on all $x_{<a_{t}}$ and
$a_{\leq t}$ that preceded it. We use $\theta$ to denote all parameters in the
PSN, but note that the factors in~\eqref{surr-model} typically only use a subset
of these. The PSN is trained to be close to $p(\xxx,\aaa)$ in terms of the
KL-divergence,
\begin{align}
  \mL(\theta) & = \kl{p(\xxx,\aaa)}{s_\theta(\xxx,\aaa)} \nonumber \\
  & = -\meanp{p(\xxx,\aaa)}{\log s_\theta(\xxx,\aaa)} + \mr{const}\eqlab{surr-kl},
\end{align}
$\mL(\theta)$ is minimized by stochastic gradient descent, which requires
calculating the unbiased gradient
estimator
\begin{equation}
\label{eq:grad_estimate}
\grad_\theta\mL(\theta)\approx-\frac{1}{N}\sum_{n=1}^N\grad_\theta\log s_\theta(\xxx^n,\aaa^n),
\end{equation}
with $(\xxx^n,\aaa^n)\overset{\mr{iid}}{\sim} p(\xxx,\aaa)$. It is crucial to
distinguish between sampling repeatedly from an empirical distribution
$\hat{p}(\xxx,\aaa)\approx p(\xxx,\aaa)$ (i.e. using a dataset) or sampling
repeatedly from $p(\xxx,\aaa)$ (online training) when calculating the gradient
\cref{eq:grad_estimate}. Either approach puts different requirements on how
$s_\theta(\xxx,\aaa)$ can be constructed. In the former case, a naive but
straightforward approach is to construct the possible address distributions and
transitions by enumerating all traces in the dataset. However, this results in a
surrogate model which is an under-approximation by construction. Furthermore, if
new data containing unseen traces is later added to the dataset the surrogate
must be reconstructed and retrained, which is computationally wasteful. Online
training, on the other hand, requires the surrogate to grow dynamically as new
data is sampled from the simulator. Our PSNs are designed to operate in the
latter case and models the space of an unbounded number of random variables. Our
method allows the PSN to grow dynamically as new traces are drawn from the
simulator yet does not require the PSN to be retrained. This is due to PSNs
having the property of being \textit{measure preserving} with respect to a set
of events $\mathcal{B}$ of particular interest as defined in
\cref{def:preserving}.

\begin{definition}\label{def:preserving} Consider a probability space
  $\paren{\Omega, \mathcal{F}, \mathbb{P}^{g}}$ and let $g\in\mathcal{G}$ be a
  function parameterizing the probability measure $\mathbb{P}^{g}$. Let
  $h:\mathcal{G}\rightarrow\mathcal{G}$ be a functional mapping such that
  $\mathbb{P}^{h(g)}$ is a probability measure associated with the probability
  space $\paren{\Omega, \mathcal{F}, \mathbb{P}^{h(g)}}$. Let
  $\mathcal{B}\subseteq\mathcal{F}$ be a set of subsets. We then say that $h$ is
  measure preserving with respect to $\mathcal{B}$ if,
  \begin{equation*}
    \mathbb{P}^{h(g)}(E)=\mathbb{P}^{g}(E), \forall E\in \mathcal{B}
  \end{equation*}
\end{definition}

We will show how to choose $\mathcal{B}$ such that the PSN grows
to model newly encountered address transitions, while leaving the probability
mass placed on known address transitions invariant to this expansion. Our method is inherently designed to work in the online setting but may also be
used with an empirical trace distribution. The key to our method lies in how we
model the probability measure associated with the set of infinitely many
possible address transitions $a_{t+1}$ from address $a_{t}$. We accomplish this
by parameterizing the probability measure in a way that dynamically allows
``breaking'' the probability measure into smaller pieces. Specifically, we
consider the probability space
$\paren{\Omega_{a_{t}}, \mathcal{F}_{a_{t}},\mathbb{P}_{a_{t}}^{\zeta}}$, where
$\Omega_{a_{t}}$ is the set of possible addresses the program can transition to
from address $a_{t}$, $\mathcal{F}_{a_{t}}$ the $\sigma$-algebra, and
$\mathbb{P}_{a_{t}}^{\zeta}$ the probability measure parameterized by a neural
network $\zeta_{a_{t}}(x_{\leq a_{t}},a_{\leq t};\theta)$. We can without loss
of generality partition $\Omega_{a_{t}}$ into transitions we are certain exist,
$\mathcal{C}_{a_{t}}$, and transitions we are \textit{uncertain} about,
$\mathcal{U}_{a_{t}}$. We have that
$\Omega_{a_{t}}=\mathcal{C}_{a_{t}}\cup\mathcal{U}_{a_{t}}$ and
$\mathcal{C}_{a_{t}}\cap\mathcal{U}_{a_{t}}=\emptyset$. In practice
$\mathcal{C}_{a_{t}}$ contains transitions observed during training and grows as
we train the surrogate, while $\mathcal{U}_{a_{t}}$ contains transitions not yet
encountered. We denote the size of known address transitions as
$C=\abs{\mathcal{C}_{a_{t}}}$, and define the neural network as a mapping
$\zeta_{a_{t}}: \mathcal{A}_{<a_{t}}\times \mathcal{X}_{<a_{t}} \rightarrow \real^{C+1}$,
where $a_{\leq t}\in \mathcal{A}_{\leq a_{t}}$ and
$x_{\leq a_{t}}\in\mathcal{X}_{\leq a_{t}}$. From this we finally define the
parameterized probability measure,
\begin{equation}\label{eq:measure}
  \mathbb{P}_{a_{t}}^{\zeta}(E) = \frac{1}{Z} \begin{cases}
    e^{\zeta_{\gamma(c)}}, &\hspace{0.5em} \text{if}~E=\tub{c}~\text{and}~
    c\in\mathcal{C}_{a_{t}}\\
    e^{\zeta_{C+1}} &\hspace{0.5em}\text{if}~E=\mathcal{U}_{a_{t}},
  \end{cases}
\end{equation}
where $\gamma:\mathcal{C}_{a_{t}}\rightarrow\tub{1,\dots,C}$ is a mapping from
observed addresses to a unique ``address index'', $\zeta_{i}$ is the $i$th
output of $\zeta_{a_{t}}(x_{\leq a_{t}},a_{\leq t};\theta)$, and
$Z= e^{\zeta_{C+1}} + \sum_{ c \in \mathcal{C}_{a_{t}}}e^{\zeta_{\gamma(c)}}$ is
the normalization constant. Looking at \eqref{measure}, we see that the
probability measure can be modeled using $\zeta_{a_{t}}$ in conjunction with the
\texttt{softmax} function,
$\mathbb{P}_{a_{t}}^{\zeta}=\mr{\texttt{softmax}}(\zeta_{a_{t}}(x_{\leq a_{t}},a_{\leq t};\theta))$.

By modeling $\mathbb{P}_{a_{t}}^{\zeta}$ according to \eqref{measure} we can
consider the model to be a classifier which assigns probability to each address
transition we know exists, while also assigning probability to yet unseen
transitions. In order to relate \eqref{measure} to the address transitions
defined in \eqref{surr-model} we note that $\forall c\in \mathcal{C}_{a_{t}}$ we
can define
$s(a_{t+1}=c|\zeta_{a_{t}}(x_{\leq a_{t}},a_{\leq t};\theta))=\mathbb{P}_{a_{t}}^{\zeta}(\tub{c})$.
In a similar fashion we can $\forall u\in \mathcal{U}_{a_{t}}$ implicitly define
$s(a_t=u|\zeta_{a_{t}}(x_{\leq a_{t}},a_{\leq t};\theta))$ in terms of
$\mathbb{P}_{a_{t}}^{\zeta}(\mathcal{U}_{a_{t}})=\sum_{u\in \mathcal{U}_{a_{t}}}\mathbb{P}_{a_{t}}^{\zeta}(\tub{u})$,
where the summation is justified as the set of all addresses (and therefore
$\mathcal{U}_{a_{t}}$) is countably infinite.
The address transition probability, $s(a_t=u|\zeta_{a_{t}}(x_{\leq a_{t}},a_{\leq t};\theta))$, would be one of the
terms in the sum. To provide some intuition on how to use the parameterized
probability measure in \eqref{measure}, we now describe how our PSNs grows
during the optimization procedure. For every set of samples of size $N$ used to
calculate the gradient estimator (i.e. a mini-batch), enumerate all the addresses
and their transitions. For each address consider all new address transitions,
which are transitions \textit{not} found in $\mathcal{C}_{a_{t}}$. Let the set
of newly encountered address transitions be denoted $\mathcal{K}_{a_{t}}$ and
its size be denoted $K=\abs{\mathcal{K}_{a_{t}}}\leq N$. We then expand the
neural network $\zeta_{a_{t}}$ and refer to the expansion as
$\tilde{\zeta}_{a_{t}}: \mathcal{A}_{<a_{t}}\times \mathcal{X}_{<a_{t}} \rightarrow \real^{C+K+1}$.
The expansion, $\tilde{\zeta}_{a_{t}}$, has its own learnable parameters that are derived
directly from $\zeta$ and parameterizes a new probability measure
$\mathbb{P}_{a_{t}}^{\tilde{\zeta}}$. We carry out the expansion so that
$\mathbb{P}_{a_{t}}^{\tilde{\zeta}}$ is given by,
\begin{equation}
  \label{eq:new-probs}
  \mathbb{P}_{a_{t}}^{\tilde{\zeta}}(E) = \frac{1}{\tilde{Z}} \begin{cases}
    &e^{\tilde{\zeta}_{\tilde{\gamma}(c)}},~\text{if}~E=\tub{c}
    \text{and}~c \in \mathcal{C}_{a_{t}}\cup\mathcal{K}_{a_{t}}\\[1em]
    &e^{\tilde{\zeta}_{C+K+1}},~\text{if}~E=\tilde{\mathcal{U}}_{a_{t}}
  \end{cases}
\end{equation}
where
$\tilde{\mathcal{U}}_{a_{t}}=\mathcal{U}_{a_{t}}\setminus\mathcal{K}_{a_{t}}$,
$\tilde{Z}$ the new normalization constant and
$\tilde{\gamma}:\mathcal{C}_{a_{t}}\times \mathcal{K}_{a_{t}}\rightarrow\tub{1,\dots,C,C+1,\cdots, C+K}$
the new index mapping, which is equal to $\gamma$ for the same addresses already
in $C_{a_{t}}$. Specifically we have $e^{\tilde{\zeta}_{\tilde{\gamma}(c)}}=e^{\zeta_{\gamma(c)}}$ if $c \in \mathcal{C}_{a_{t}}$, $e^{\tilde{\zeta}_{\tilde{\gamma}(c)}}=e^{\zeta_{C+1} - \log\paren{K+1}}$ if $c\in\mathcal{K}_{a_{t}}$, and $e^{\tilde{\zeta}_{C+K+1}} = e^{\zeta_{C+1} - \log\paren{K+1}}$. This choice, \eqref{new-probs}, leads to the following theorem, which we prove in \appref{proof-1},

\begin{theorem} \label{theorem:1} Consider a probability measure
  $\mathbb{P}_{a_{t}}^{\zeta}$ characterized by a neural network
  $\zeta_{a_{t}}\in\mathcal{G}$ according to \eqref{measure}. Consider also a
  sample of traces of size $N$ which for each address $a_{t}$ contain a set
  $\mathcal{K}_{a_{t}}$ of new address transitions. Let the expansion procedure
  represented by \cref{eq:new-probs} be defined as the function
  $h:\mathcal{G}\rightarrow\mathcal{G}$ such that
  $\tilde{\zeta}_{a_{t}}=h(\zeta_{a_{t}})$. If
  $\mathcal{B}=2^{\mathcal{C}_{a_{t}}}\cup\tub{\mathcal{U}_{a_{t}}}\subseteq\mathcal{F}_{a_{t}}$
  where $2^{\mathcal{C}_{a_{t}}}$ denotes the powerset of $\mathcal{C}_{a_{t}}$, then
  for all addresses $a_{t}$, the functional mapping $h$ is measure preserving
  with respect to $\mathcal{B}$ as defined in \cref{def:preserving}, and
  \begin{equation*}
  \mathbb{P}^{\tilde{\zeta}}_{a_{t}}(E)=\mathbb{P}^{\zeta}_{a_{t}}(E), \forall E\in\mathcal{B}
  \end{equation*}
\end{theorem}

Once a new probability measure is created, the new transitions found in
$\mathcal{K}_{a_{t}}$ are added to $\mathcal{C}_{a_{t}}$. In \cref{fig:expansion} we show an illustration of the expansion process and we provide further
details on how to expand the PSNs when encountering new address transitions in
\appref{alg} where we also provide \cref{algo:psn-expand}. Following the
expansion of the PSN, the update of the PSN parameters, $\theta$, is carried out
by calculating the gradient estimator and performing gradient descent. This
procedure is repeated until convergence. Additional details and design choices
of PSNs can be found in \cref{app:psn-design}.

\subsection{Evaluating and Executing PSNs}
\label{sec:eval}

The construction of our PSNs described above ensures that the surrogate models
define a probability measure on spaces with an unbounded number of random
variables. In particular, and we prove this in~\cref{app:proof-2},

\begin{theorem}\label{theorem:any-evaluation}
  Let $s(\xxx,\aaa)$ be a surrogate model using PSNs. Then any trace
  $(\xxx,\aaa)\sim p(\xxx,\aaa)$ can be evaluated under $s(\xxx,\aaa)$.
\end{theorem}

While \cref{theorem:any-evaluation} guarantees evaluation for all possible
traces generated by the reference simulator, the surrogate $s$ is only likely to
provide accurate density estimates for traces for which all addresses have been
encountered during training. As such, at evaluation time when training is
complete, it is of more practical use to place zero probability measure on
traces containing unknown addresses. The justification of this choice becomes
more apparent when discussing the execution of PSN-based surrogate models. Such
executions start with the \texttt{begin-execution} address, after which the
surrogate samples a new address from the transition distribution, a value is
sampled from the distribution at the sampled address, after which a new address
transition is sampled, \emph{etcetera}, finishing only when the surrogate
samples an \texttt{end-execution} address. The procedure is illustrated in more
detail in \cref{fig:construct} in \cref{app:psn-design}. The question now arises
what should happen if the surrogate samples an \texttt{unknown} address at any
point during its execution. Recall that at each address $a_{t}$ the probability
associated with such an event is
$\mathbb{P}^{\zeta}_{a_{t}}(\mathcal{U}_{a_{t}})$. One straightforward approach
would be to (1) Generate a new arbitrary address including the possibility to
generate an \texttt{end-execution} address. (2) If the new address is not an
\texttt{end-execution} address then expand the PSN according to
\cref{eq:new-probs} in order to accommodate the newly generated address. (3)
Sample some distribution from a prior distribution over distributions. (4)
Repeat until an \texttt{end-execution} address is generated. Clearly, the produced traces
from such a procedure will almost certainly have zero probability under the
reference simulator, and would yield spurious results. To remedy this, we
instead decide to only allow transitions between addresses encountered during
training. Specifically, whenever an \texttt{unknown} address is sampled, we keep
resampling until a known address is sampled, leading to the following adjusted
address transition probabilities for all $c\in\mathcal{C}_{a_{t}}$,
\begin{figure}[t!]
  \centering
  \includegraphics[scale=0.5]{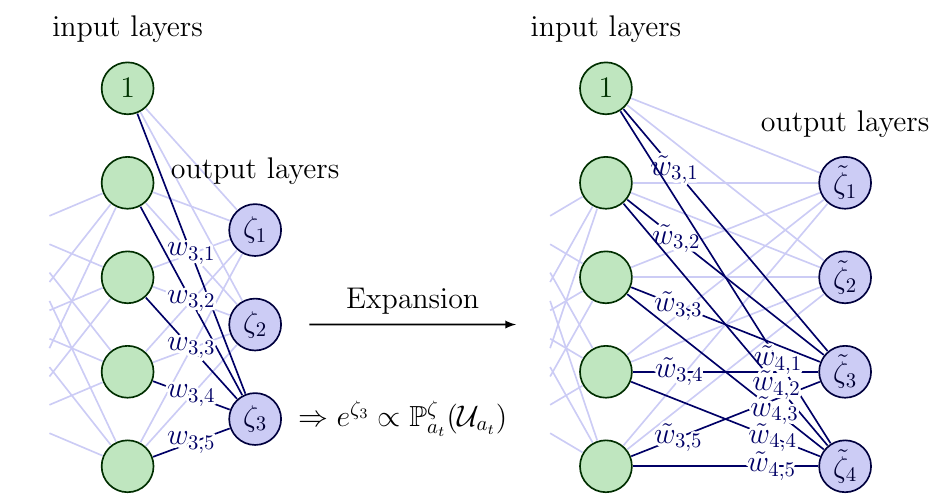}
  \caption{Illustration of the PSN expansion relating \cref{eq:measure} and
  \cref{eq:new-probs}. The expansion takes place in the final address transition
  prediction layer, and the new weights $\tilde{w}_{\cdot,\cdot}$ relate to the
  former weights $w_{\cdot,\cdot}$ as follows: (1) for the weights associated
  with known address transitions we have $\tilde{w}_{i,j}=w_{i,j}$ for all
  $i,j\in \tub{1,2}\times\tub{1,\dots,5}$. (2) For the weights associated with
  the unknown addresses and newly encountered addresses we have for all
  $i,j\in\tub{3,4}\times\tub{2,\dots,5}$ that $\tilde{w}_{i,j}= w_{3,j}$ and for
  the bias weights $\tilde{w}_{i,1} = w_{3,1} - \log(K+1)$. In this case, with
  one newly encountered address, we have $K=1$.}
  \figlab{expansion}
\end{figure}
\begin{align}\label{eq:rejection-transition}
  P(a_{t+1}=c|A) & = \frac{P(A|c)s(c|\zeta_{a_{t}}(x_{\leq a_{t}},a_{\leq t};\theta))}{P(A)} \nonumber \\
  &= \frac{s(c|\zeta_{a_{t}}(x_{\leq a_{t}},a_{\leq t};\theta))}{1-\mathbb{P}_{a_{t}}^{\zeta}(\mathcal{U}_{a_{t}})},
\end{align}
where $A$ denotes the event that we accept the address transition
$a_{t}\rightarrow a_{t+1}$.
\begin{figure*}[t!]
  \centering
  \includegraphics[scale=0.60]{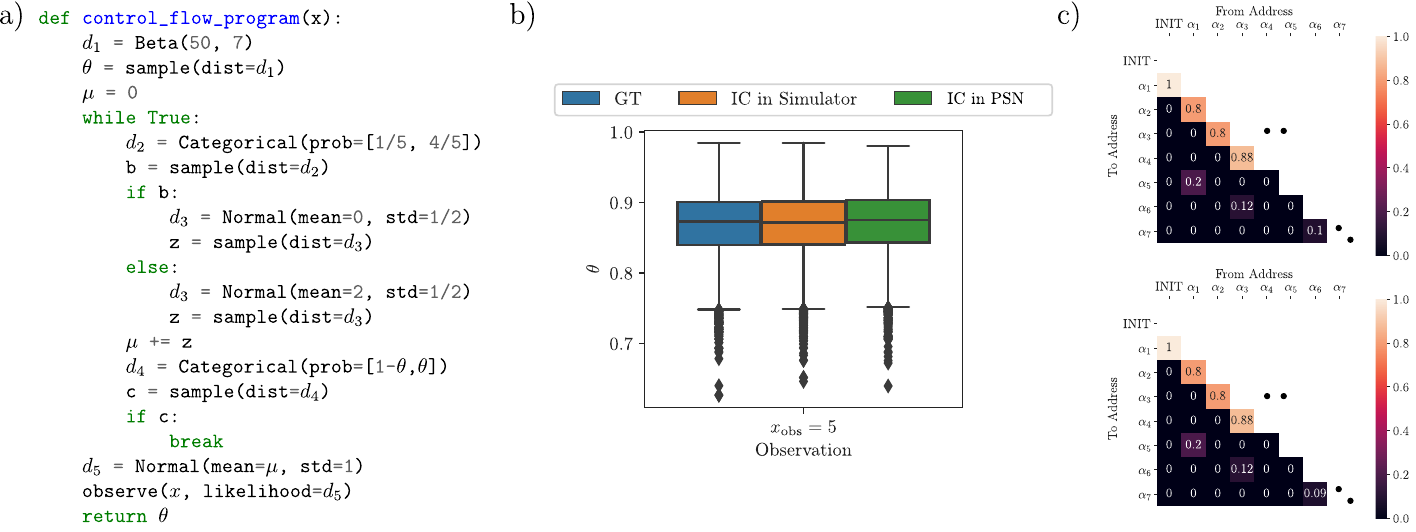}
  \caption{(a) Program containing stochastic control flow in the form of a
    for-loop with a nested if-else statement. The task here is to perform
    posterior inference about $\theta$ given the observed value of $x$. (b) Each
    boxplot represents the estimated posterior distribution of $\theta$
    conditioned on $x=5$. We see that inference using IC in the simulator and
    using IC in the PSN both are identical to GT. (c) shows a subset of the
    address transitions with high probability for the original program (top) and
    our PSN (bottom). We observe identical transition probabilities which,
    together with (b), shows that the PSN is able to approximate models with
    complex control flow.}
  \figlab{prog-pos}
\end{figure*}
\subsection{Practical Limitations}
\label{sec:limitations}
While PSNs target programs written in universal PPLs, there are practical
considerations accompanying (1) the rejection sampling step of address
transitions and (2) the proposed use of RNNs as the core of the PSN. Regarding
(1), the rejection sampling step is equivalent to placing zero probability mass
on traces that were not observed during training. In general this results in the
adjusted address transition probabilities, \cref{eq:rejection-transition}, to
become slightly biased. Additionally, this implies that PSNs become an
under-approximation of the target simulator, which may have non-zero probability
on certain traces where the PSN places no probability mass. In the limit of
observing all possible address transitions these issues simply vanish, while
practically the more traces are observed, the less likely it becomes that
important addresses with high probability mass are missed. Concerning (2), the
choice of using RNNs to model the flow of information (i.e. inter-variable
dependencies) was made as it has proven very effective in practice.
Notwithstanding, while RNNs are capable, in theory, of emulating Turing
machines~\citep{weiss2018practical,siegelmann1998neural,siegelmann1994analog,siegelmann1995computational,chen2018recurrent},
finite memory and floating point precision make them finite state machines in
practice. For target programs, which require storing information on e.g. a
potentially infinitely growing stack, we would not, in general, expect RNNs to
model said programs arbitrarily well. This does not, however, influence the
results in \cref{theorem:1,theorem:any-evaluation} which are agnostic to the
specific implementation of the dependency model. Rather, it implies that the
size of the RNN needs to be chosen appropriately to ensure that accurate
surrogates are learned. Put in different words, the approximating distribution
has limited flexibility, but its support is guaranteed to be correct by
\cref{theorem:any-evaluation}. If RNNs turn out insufficient, we suggest
considering differentiable neural computers~\citep{graves2016hybrid} as a
potential suitable alternative, as it has access to external memory.
Furthermore, previous work by \citet{harvey2019attention} suggests that the
transformer architecture \citep{vaswani2017attention} might also be a good
alternative choice in some cases.

\subsection{Complexity Analysis}

We limit the complexity analysis to pertain to the number of addresses
encountered during training - a set we denote $A$. We start by considering the worst-case scenario, where the possible addresses
transitions of some program is as follows: Order all addresses on a single line
in the order in which they could appear. If a transition can occur from any
address to any other address following it, the computational complexity must be
$\mO(A^{2})$. This is also true memory-wise. This is because from any particular
address we must calculate the transition probability to any of the addresses
following it. This includes storing model parameters to each of those potential
addresses.

\begin{figure*}[t!]
  \centering
  \includegraphics[scale=0.65]{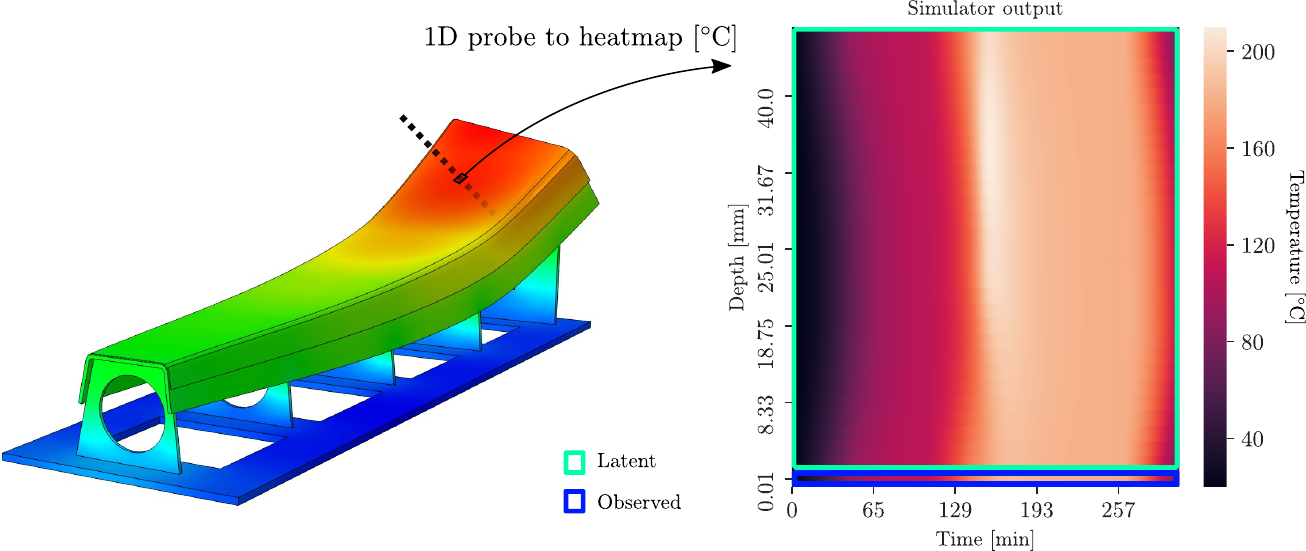}
  \caption{Composite manufacturing involves an uncured composite material being
    laid up onto a tool, of known material, which are then placed in an
    autoclave where a predefined pressure and heating cycle is imposed (left).
    We consider a 1D simulation of this process as a function of time, leading
    to the 2D heatmap (right). The set of latent variables are heat transfer coefficients
    and thicknesses and internal temperature (green box). The observed variables
    are temperature configuration of the autoclave, air temperatures, and tool
    temperature (blue box, measured at the bottom surface of the tool).}
  \label{fig:raven-prob}
\end{figure*}

How the PSNs compare to the reference simulator complexity-wise cannot generally
be determined. We imagine that the reference simulator in many cases has similar
complexity. For instance, if the various address transitions are due to
\texttt{if}-\texttt{elif}-\texttt{else} statements in a program, the reference simulator may calculate all logical clauses leading to complexity $\mO(A^{2})$. However, there might exist an equivalent program which is much more efficient in how it determines its state
transitions, possibly even $\mO(A)$, but we cannot in general make such
guarantees. Similarly, there may exist other programs which scale much worse,
say $\mO(x^{A})$, $x>1$ - we can imagine programs which do complex computations
that reason about all possible future and past states.

Ultimately, what matters in determining whether or not to use a PSN to replace
the reference simulator is the wall-clock time of the PSN versus the reference
simulator.

\section{Experiments}
\label{sec:experiments}

\subsection{Stochastic Control Flow}
Here we present an experiment that highlights the PSN's capability to learn a
model's address transitions. \cref{fig:prog-pos}(a) shows a program with complex
stochastic control flow, where the aim is to perform posterior inference of
$\theta$ given the observed value of $x=5$ using a trained PSN.
\cref{fig:prog-pos}(b) shows boxplots representing the estimated posterior
distribution of $\theta$ conditioned on $x=5$. The inference results are
obtained using either MCMC, specifically Lightweight Metropolis-Hastings
(LMH)~\citep{wingate2011a}, with a chain length of 1,000,000 samples (denoted GT
for ground truth) or IC in either the simulator or PSN using 10,000 resampled
importance weighted samples. To evaluate the address transition capability we
look at \cref{fig:prog-pos}(c), which shows a subset of address transitions with
high probability observed across 50,000 generated samples from the model (top)
and PSN (bottom). We observe that the three posteriors and the address
transition probabilities are identical. Together these results show that the PSN
has successfully approximated the program including the address transitions
associated with the original program. Further evidence can be found
in~\cref{app:model_spec_flow,app:transitions}.
\subsection{Program synthesis}
Next we consider the question of whether the machinery
as it is presented here is able to capture relevant connections between the
addresses as they are available. As touched upon in~\cref{sec:limitations}, RNNs
are finite state machines in practice and so may be insufficient in accurately
modeling the inter-variable dependencies. To shed light on this we provide an
experiment that showcases that RNNs can, in practice, model programs which
require access to dynamically growing memory. Particularly, we learn a surrogate
for a model that generates valid Python programs. We use a subset of the Python
syntax that allows \texttt{if}, \texttt{else}, and \texttt{for} statements, to
an unbounded nesting depth, corresponding to piecewise linear functions. Example
programs and full technical details of the simulator can be found in
\cref{app:prog-synthesis-details}. The crucial element of this experiment is the
existence of a stack in the original simulator, that tracks the opening and
closing of conditionals, and determines at any time what constitutes a valid
next line. The surrogate has to store this information in the RNN hidden state,
or alternatively, learn the valid continuations that belong to a certain
unbounded collection of addresses. We judge the
quality of PSNs by the fraction of valid programs that are generated.
As the validity of programs allows for direct evaluation without performing
inference, we omit the latter. We find the percentage of valid programs to be
99.62\% (50k samples). We thus conclude that in practice, the use of an RNN for
our method is easily sufficient for a task requiring the simulation of a program
stack.

\subsection{Process Simulation of Composite Materials}
\label{sec:raven}
\begin{figure}[t!]
  \centering
  \includegraphics[scale=0.67]{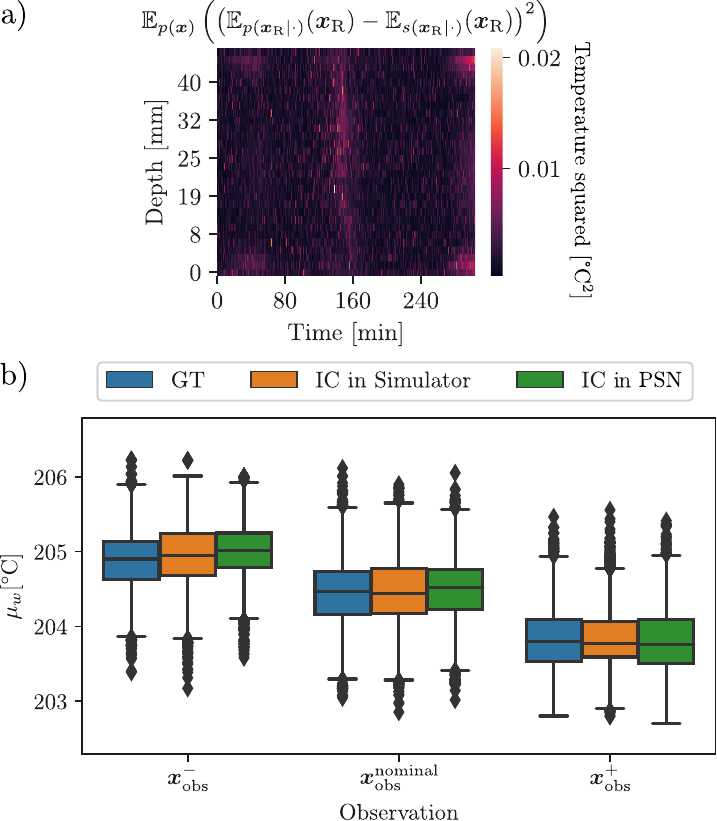}
  \caption{(a) Shows the expected squared difference between the output
    ($\xxx_{\mr{R}}=\xxx_{\mr{RAVEN}}$) from the PSN and output from the
    simulator. We observe negligible errors throughout, with small peaks around
    time $\SI{160}{\min}$ and towards the end of the heating process at the
    top/bottom. Each boxplot in (b) represents the estimated posterior
    distribution (conditioned on either $\xxx^-_{\mr{obs}}$,
    $\xxx^{\mr{nominal}}_{\mr{obs}}$, or $\xxx^+_{\mr{obs}}$) over a fixed time
    window $f(\xxx) = \mu_w$. We see that IC in the PSN yields effectively the
    same posterior as IC in the simulator across all observations. In all cases
    inference using IC agrees with the ground truth (GT). }
  \figlab{posterior}
\end{figure}
In this experiment we train a surrogate model for a commercial heat-transfer
finite element analysis simulator, depicted in \cref{fig:raven-prob}, that is
used to model the cure cycle for composite aircraft (e.g. Boeing) parts. We show
how to use inference to estimate the temperature of the part in regions that
cannot be accessed non-invasively. Such results are critical for determining whether
the part is safe or not. The particular simulator used is RAVEN which simulate
the curing process of composite materials, a proprietary software developed
by~\citet{Technologies2019}. RAVEN is used in the aerospace and automotive
industries to evaluate key performance metrics for part manufacturing design
with the ultimate goal of decreasing manufacturing cost whilst retaining part
performance and safety. Physical observations of the material's internal
temperature during manufacturing are expensive, if not impossible, and
manufacturers would prefer to infer the internal state of the material given
less expensive external measurements. \cref{fig:raven-prob} illustrates this
process and the experimental setup. Using probabilistic programming and our
PSNs, we seek to infer the internal state of the material conditioned on
realistically observable quantities. We evaluate the quality of the PSN by
considering the expectation of $\xxx_{\mr{RAVEN}}$ (material temperature during
processing) conditioned on the RAVEN configurations, $\xxx_{\mr{config}}$, under
(1) the model distribution
$\meanp{p(\xxx_{\mr{RAVEN}}|\xxx_{\mr{config}})}{\xxx_{\mr{RAVEN}}}$ and (2) the
PSN distribution
$\meanp{s(\xxx_{\mr{RAVEN}}|\xxx_{\mr{config}})}{\xxx_{\mr{RAVEN}}}$.
~\cref{fig:posterior}(a) shows generally negligible expected squared errors
between the surrogate and simulator outputs, and we provide additional results
in \appref{raven} that showcase the efficacy of the trained PSN. Small peaks
are, however, observed in~\cref{fig:posterior}(a) at around time
$t=\SI{160}{\min}$, as well as towards the end of the heating process, which is
where the internal temperature exhibits the most rapid changes.

To evaluate the quality of performing inference (using IC) in the PSN we
consider the scenario where we only observe the configurations, air and surface
temperatures of the curing process, $\xobs$. The latent variables $\xlat$ are
the dimensions of the material, the heat transfer coefficients and the internal
temperature during curing. We then consider the function
$f(\xxx_{\mr{lat}}) = \mu_w$, being the empirical mean of the internal
temperature of the material across the time window
$w=\br{\SI{155}{\minute},\SI{165}{\minute}}$ (chosen to be close to peak
temperatures) and at a fixed depth $\SI{30}{\mm}$ (chosen to be somewhere near
the upper quarter of the material). We then estimate
$\hat{\mu}_w\approx\meanp{p(\xxx_{\mr{lat}}|\xxx_{\mr{obs}})}{f(\xxx_{\mr{lat}})}=\meanp{p(\xxx_{\mr{lat}}|\xxx_{\mr{obs}})}{\mu_w}$
using IC with the same inference network $q(\xxx_{\mr{lat}}|\xxx_{\mr{obs}})$
used for performing inference in both the surrogate and the model. As a ground
truth posterior, we employ SIS where the proposal distribution is the prior
$q(\xxx_{\mr{lat}}|\xxx_{\mr{obs}})=p(\xxx_{\mr{lat}})$ and denote it GT. To
evaluate the effect of amortized inference we consider conditioning on three
different observations $\xxx^-_{\mr{obs}}$, $\xxx^{\mr{nominal}}_{\mr{obs}}$,
and $\xxx^+_{\mr{obs}}$ each corresponding to an observation produced by the
simulator with input values and temperature settings well below, equal to, and
well above the nominal values respectively. In all cases inference is performed
using 15,000 traces (SIS particles) and we summarize the results
in~\tabref{inference}. We show that performing inference in the PSN yields
approximately the same results as inference in the simulator. We only find small
deviations when observing $\xxx_{\mr{obs}}^-$ where the PSN seems to barely
overestimate $\hat{\mu}_w$ compared to the GT. To get a sense of how our traces
are distributed we show in~\cref{fig:posterior}(b) boxplots representing the
posterior distribution from which we estimate
$\meanp{p(\xxx_{\mr{lat}}|\xxx_{\mr{obs}})}{\mu_w}$. Each boxplot is made by
resampling the 15,000 importance weighted samples. The results confirm that
inference in the PSN yields similar posteriors compared to inference in the
simulator. These boxplots also illustrate why $\hat{\mu}_w$ was slightly
overestimated when doing inference in the PSN; when observing
$\xxx_{\mr{obs}}^-$ the posterior is shifted slightly upwards compared to the
GT.

\begin{table}[t!]
  \centering
  \caption{ We estimate
    $\hat{\mu}_w\approx\meanp{p(\xxx_{\mr{lat}}|\xxx_{\mr{obs}})}{\mu_w}$ under
    the posterior using SIS denoted GT, IC in simulator (ICS) and IC in PSN
    (ICP) using 15,000 traces, and report the associated effective sample size
    (ESS). We provide six different estimates; three posteriors and three
    observations $\xxx^-_{\mr{obs}}$, $\xxx^{\mr{nominal}}_{\mr{obs}}$, and
    $\xxx^+_{\mr{obs}}$. We observe that the PSN estimates matches that of the
    GT and IC in the simulator.}
  \begin{tabular}{lllllll}
    \toprule
    & \multicolumn{2}{c}{$\xxx_{\mr{obs}}^-$} & \multicolumn{2}{c}{$\xxx_{\mr{obs}}^{\mr{nominal}}$} & \multicolumn{2}{c}{$\xxx_{\mr{obs}}^+$}\\
    \midrule
    & $\hat{\mu}_w$ & ESS & $\hat{\mu}_w$ & ESS & $\hat{\mu}_w$ & ESS \\
    \midrule
    GT & 204.90 & 259 & 204.46 & 304 & 203.82 & 399 \\
    ICS & 204.96 & 158 & 204.46 & 340 & 203.83 & 204 \\
    ICP & 205.01 & 173 & 204.49 & 279 & 203.80 & 292 \\
    \bottomrule
  \end{tabular}
  \tablab{inference}
\end{table}

The advantage of using the PSN is that we maintain high accuracy in the
posterior estimates with a speedup factor of 15.32 when comparing the number of
traces generated per second. Furthermore, in cases where we simply seek to produce
faster simulations (not for the sake of inference), the PSN provides an even
greater speedup factor of 90.16. The additional speedup is due to dropping the
overhead of performing inference. The exact running times and model
specifications can be found in~\cref{app:running_times,app:model_spec_raven}.

\section{Related Work}

As far as the authors of this paper are aware, the PSN is the first framework
for learning surrogate models that models simulators containing a potentially
unbounded number of random variables by automatically extracting and using a
simulator's latent structure. Surrogate modeling is, however, a topic that dates
back several decades and is fundamentally a regression problem, where the
surrogate predicts the output of the model for a given input. Currently, the
most commonly used methods for constructing deterministic surrogate
models~\citep{razavi2012a} include Kriging~\citep{simpson2001a,
  sacks1989design}, support vector machines (SVMs)~\citep{willcox2005fourier},
radial basis functions (RBFs)~\citep{hussain2002metamodeling,
  mullur2006metamodeling}, and neural networks (NNs)~\citep{tompson2017a,
  khu2003reduction, gilmer2017a}, while methods like the stochastic
Kriging~\citep{hamdia2017stochastic} allow for stochastic surrogate modeling.
Notwithstanding, such commonly used methods are incompatible with simulators
with an unbounded number of variables.

Finally, the idea of learning trace executions using LSTMs has been studied
before, see for example neural programmer-interpreters
(NPI)~\citep{reed2015neural}. Methods like NPIs are trained to predict the
sequence of called subroutines used to solve specific tasks like sorting or
image rotation. As such, NPIs make no attempt to abstract away the predicted
subroutines. That is, if any subroutine causes a computational bottleneck, NPIs
cannot decrease the computational cost. This is fundamentally different to our
PSN surrogate method which aims to model the entire simulator.

\section{Conclusions}
% \label{sec:conclusions}

We have proposed \textit{probabilistic surrogate networks}, a novel approach to
surrogate modeling that considers not only the distributions in stochastic
simulators but the stochastic structure of the simulator itself. Our main
contribution is to develop a construction in which the surrogates allow for the
description of a dynamically growing number of random variables, while
maintaining consistency of the assigned probability measure as new variables are
encountered. Such a framework is a requirement for producing surrogates for
arbitrary simulators potentially containing an unbounded number random
variables. Using a real-world process simulation of composite materials as an
example, we have shown that our approach provides significant computational
speedup in inference problems using inference compilation, while preserving the
quality of inference results that are indistinguishable from the
ground truth.

\begin{acknowledgements} % will be removed in pdf for initial submission,
                         % so you can already fill it to test with the
                         % ‘accepted’ class option
  We acknowledge the support of the Natural Sciences and Engineering Research
  Council of Canada (NSERC), the Canada CIFAR AI Chairs Program, and the Intel
  Parallel Computing Centers program. Additional support was provided by UBC's
  Composites Research Network (CRN), and Data Science Institute (DSI). This
  research was enabled in part by technical support and computational resources
  provided by WestGrid (www.westgrid.ca), Compute Canada (www.computecanada.ca),
  and Advanced Research Computing at the University of British Columbia
  (arc.ubc.ca).

\end{acknowledgements}

\bibliography{bibtex.bib}

\appendix
\onecolumn
\section{Proofs}
\subsection{Proof of Theorem~\ref{theorem:1}}
\label{app:proof-1}

For an address $a$ define $\mathcal{C}$ and $\mathcal{K}$ as specified in
\cref{sec:psn}. That is $\mathcal{C}$ is the set of address transitions we know
are possible and $\mathcal{K}$ is the set of newly encountered address
transitions found in a sample of traces drawn from a reference simulator. Let
$C=\abs{\mathcal{C}}$ and $K=\abs{\mathcal{K}}$ be the size of each set
respectively. We consider the set of previous unknown address transitions
$\mathcal{U}$ and denote the new set of unknown transitions
$\tilde{\mathcal{U}}=\mathcal{U}\setminus \mathcal{K}$. Finally, define the
probability measures $\mathbb{P}$ and $\tilde{\mathbb{P}}$ both associated with
the sample space $\Omega$ and $\sigma$-algebra $\mathcal{F}$ according to

\begin{align*}
  \mathbb{P}(E) & = \frac{1}{Z} \begin{cases}
    e^{\vvv_{\gamma(c)}}, &\quad  \text{if}~E=\tub{c}~\text{and}~c \in \mathcal{C}\\
   e^{\vvv_{C+1}} &\quad \text{if}~E=\mathcal{U}
   \end{cases} \\
    \tilde{\mathbb{P}}(E) & = \frac{1}{\tilde{Z}} \begin{cases}
      e^{\vvv_{\gamma(c)}}, &\quad  \text{if}~E=\tub{c}~\text{and}~c \in \mathcal{C}\\
      e^{\vvv_{C+1} - \log\paren{K+1}}, &\quad  \text{if}~E=\tub{k}~\text{and}~ k\in\mathcal{K}\\
    e^{\vvv_{C+1} - \log\paren{K+1}}, &\quad \text{if}~E=\tilde{\mathcal{U}},
    \end{cases}
\end{align*}
where $\vvv\in\real^{C+1}$, $Z$ and $\tilde{Z}$ are normalization constants, and
$\gamma:\mathcal{C}\rightarrow\tub{1,\dots,C}$ is a mapping from observed
addresses to a unique ``address index''.

Observe that the relationship between $\tilde{\mathbb{P}}$ and $\mathbb{P}$ is
equivalent to the relationship between $\mathbb{P}_{a_{t}}^{\tilde{\zeta}}$ and
$\mathbb{P}_{a_{t}}^{\zeta}$ defined in \cref{sec:psn}. In particular, we
consider the functional mapping $h:\mathcal{G}\rightarrow\mathcal{G}$ such that $\tilde{\zeta}=h(\zeta)$, where
$\tilde{\zeta},\zeta\in\mathcal{G}$. The
proof of \cref{theorem:1} therefore reduces to proving that for all
$E\in\mathcal{B}=2^{\mathcal{C}}\cup\tub{\mathcal{U}}\subseteq\mathcal{F}$, $\tilde{\mathbb{P}}(E)=\mathbb{P}(E)$ holds.

We start by comparing the normalization constants:

\begin{align}
  \label{eq:proof-norm}
  \tilde{Z} &= \sum_{c\in\mathcal{C}}e^{\vvv_{\gamma(c)}} + \sum_{k\in\mathcal{K}}e^{\vvv_{C+1}-\log(K+1)} + e^{\vvv_{C+1}-\log(K+1)}\nonumber \\
            &=  \sum_{c\in\mathcal{C}}e^{\vvv_{\gamma(c)}} + (K+1)e^{\vvv_{C+1}-\log(K+1)}\nonumber \\
            &= \sum_{c\in\mathcal{C}}e^{\vvv_{\gamma(c)}} + e^{\vvv_{C+1}} \nonumber \\
            &= Z,
\end{align}

leading to,
\begin{figure*}[t!]
  \centering
  \includegraphics[scale=0.2]{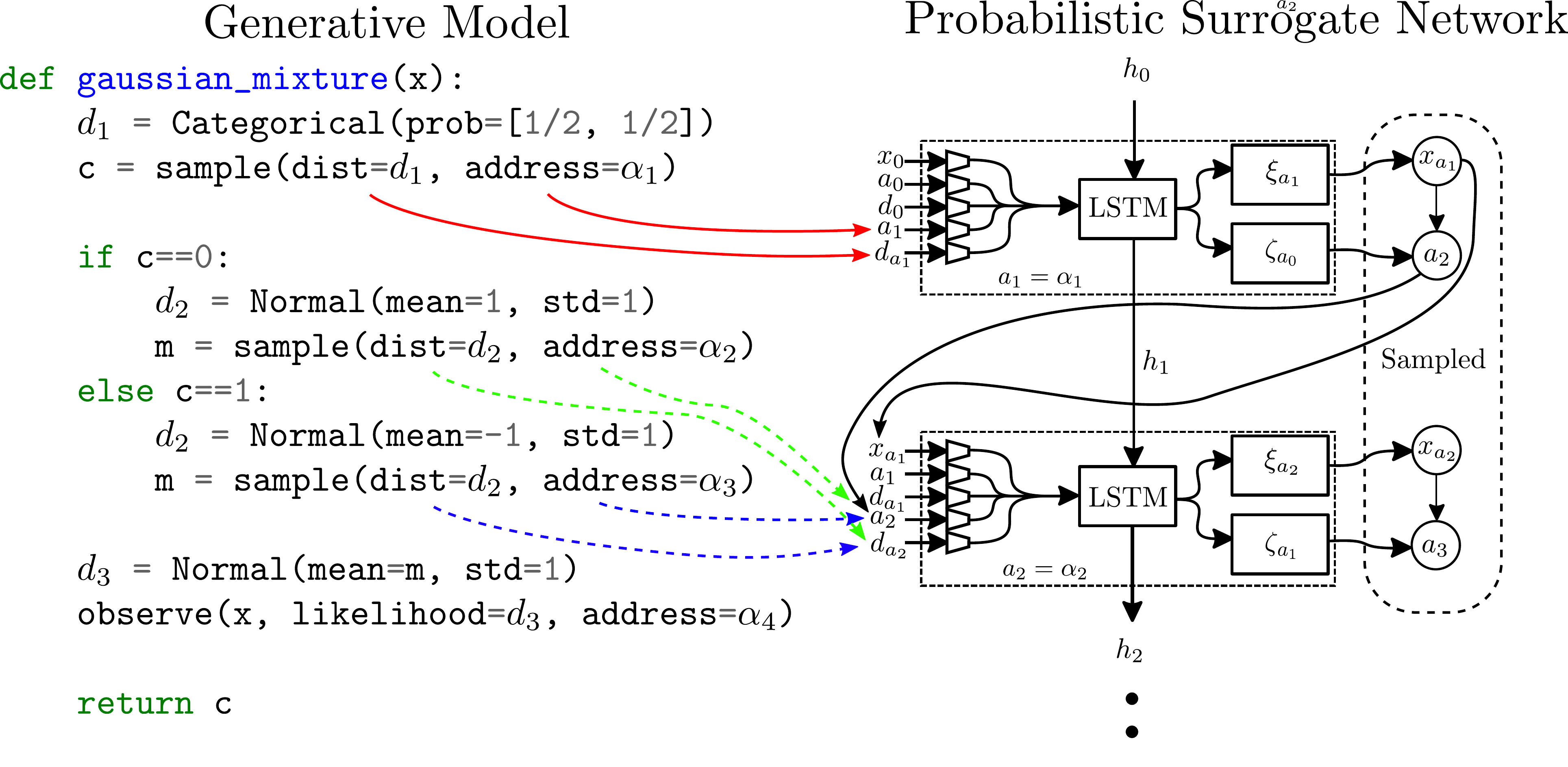}
  \caption{ Illustration of the equivalence between a simple generative model
    and a probabilistic surrogate network. The red arrows represent
    what is extracted from the program and fed to the surrogate network during
    training. Generally, this would be an address $a$ and the distribution type
    $d_{a}$ at that address. This extraction happens at every address
    encountered when executing the program while training the surrogate. The
    dashed arrows represents possible extractions after one step of running the
    PSN. Which extraction depends on the sampled value \texttt{c}. If $c=1$
    then $a_{2}=\alpha_{2}$ and the blue dashed arrow extraction happens
    otherwise $a_{2}=\alpha_{3}$ and the green dashed arrow extraction happens.}
  \figlab{construct}
\end{figure*}
\begin{align}
  \tilde{\mathbb{P}}(\{c\}) &= \frac{1}{\tilde{Z}}e^{\vvv_{\gamma(c)}} = \frac{1}{Z} e^{\vvv_{\gamma(c)}} = \mathbb{P}(\{c\}) \quad \forall c\in\mathcal{C} \label{eq:single-eq}\\
  \tilde{\mathbb{P}}(\mathcal{K}\cup \tilde{\mathcal{U}}) &= \tilde{\mathbb{P}}(\mathcal{U}) = \tilde{\mathbb{P}}(\tilde{\mathcal{U}}) + \tilde{\mathbb{P}}(\mathcal{K}) \nonumber\\
                            &= \frac{1}{\tilde{Z}}\paren{e^{\vvv_{C+1}-\log(K+1)} +\sum_{k\in\mathcal{K}}e^{\vvv_{C+1}-\log(K+1)}} \nonumber\\
  &= \frac{1}{Z}e^{\vvv_{C+1}} = \mathbb{P}(\mathcal{U}) \label{eq:unknown-eq}.
\end{align}

Since all events $\tub{\{c\}|c\in\mathcal{C}}$ are mutually exclusive, it follows
from \cref{eq:single-eq} that

\begin{equation}\label{eq:subsets-eq}
  \tilde{\mathbb{P}}(E) = \sum_{e\in E}\tilde{\mathbb{P}}(\{e\}) = \sum_{e\in E}\mathbb{P}(\{e\}) = \mathbb{P}(E), \quad \forall E\in 2^{\mathcal{C}}.
\end{equation}

Combining \cref{eq:unknown-eq} and \cref{eq:subsets-eq}, we arrive at the final result,

$$
\tilde{\mathbb{P}}(E)=\mathbb{P}(E), \quad \forall E\in\mathcal{B}=2^{\mathcal{C}}\cup\tub{\mathcal{U}},
$$

which completes the proof. \qed

\subsection{Proof of Theorem~\ref{theorem:any-evaluation}}
\label{app:proof-2}
The proof of \cref{theorem:any-evaluation} only requires the consideration of
two possible scenarios regarding a trace $(\xxx,\aaa)$: (1) the trace either
contains address transitions observed during the training of $s(\xxx,\aaa)$ in
which case its evaluation is straightforward. (2) $(\xxx,\aaa)$ contains
addresses and transitions not encountered during training. In the latter case,
we would simply expand our PSN to account for those new transitions according to
\cref{eq:new-probs}.\qed

\section{Algorithms}
\label{app:alg}
The procedure we use to expand the address transition distribution at address $a_t$ upon encountering a set of yet unseen transitions $\mathcal{K}_{a_t}$ is outlined in \cref{algo:psn-expand}. The procedure is applied to the final layer of a neural network which follows an intermediate layer of size $n_{emb}$. The operation $\mathrm{detach}(\cdot)$ denotes duplication without copying the gradient information, hence detaching the argument from the computational graph. The $\mathrm{concat}(\cdot,\cdot)$ operation concatenates the second argument to the first, and re-attaches the newly created matrix or vector to the computational graph as a leaf.
\begin{algorithm}[h]
  \DontPrintSemicolon
  \KwIn{A set $\mathcal{K}_{a_t}$ of new address transitions with size $K = \abs{\mathcal{K}_{a_t}}$}
  \KwIn{Weights $\vec{W} \in \mathbb{R}^{ (C+1) \times n_{emb}}$ and biases $\vec{b} \in  \mathbb{R}^{C+1}$, with $C = \abs{\mathcal{C}_{a_t}}$}
  % \KwOut{The largest element in the set}
  $\vec{w}^{u} = \mathrm{detach}\left(\vec{w}_{C+1}\right)$ \tcp*{$\vec{w}_{C+1}$ denotes row $C+1$ of $\vec{W}$}
  $b^{u} = \mathrm{detach}\left(b_{C+1}\right) - \log (1 + K)$ \tcp*{${b}_{C+1}$ denotes element $C+1$ of $\vec{b}$}
  $\vec{W} = \vec{W}_{:C}$ \tcp*{$\vec{W}_{:C}$ denotes the first $C$ rows of $\vec{W}$}
  $\vec{b} = \vec{b}_{:C}$ \tcp*{$\vec{b}_{:C}$ denotes the first $C$ elements of $\vec{b}$}
  \For{$k=0$ \KwTo $K+1$} {
	$\vec{W} = \mathrm{concat}(\vec{W}, \vec{w}^{u})$ \\
	$\vec{b} = \mathrm{concat}(\vec{b}, b^{u})$ \\
      }
  \caption{PSN address transitions expansion. Definitions of the $\mathrm{detach}$ and $\mathrm{concat}$ operations are given in \appref{alg}}
  \label{algo:psn-expand}
\end{algorithm}

\section{Surrogate Network Architecture}
\label{app:psn-design}

The PSN architecture is dynamically constructed during training and uses an LSTM
core as well as embeddings of the addresses, distribution types, and other
random variables. These embeddings are referred to as $a_{i}$, $d_{i}$, $x_{i}$
respectively. In particular, each address is associated with a fixed
distribution type. These deterministic and fixed pairings between addresses and
distribution types are stored and made part of the surrogate model. In other
words, when constructing the PSN we know the distribution type associated with
each address. The dynamic construction is driven by the program, where the
embeddings are fed to the LSTM core whose output is then fed to so-called
``distributions layers'' $\xi_{a_t}$ and $\zeta_{a_{t}}$, that for each unique
address $a_t$ produces the parameters for
$s(x_{a_t}|\xi_{a_t}(x_{<a_{t}},a_{\leq t},\theta))$ and
$s(a_{t+1}|\zeta_{a_{t}}(x_{<a_{t+1}},a_{\leq t},\theta))$ respectively. Note
that the value sampled from $s(x_{a_t}|\xi_{a_t}(x_{<a_{t}},a_{\leq t},\theta))$
is additionally fed to $\zeta_{a_{t}}$. In practice, this means that all
conditional probabilities of the PSN are conditioned on the distribution types
and therefore their embeddings $d_{i}$. While not part of the problem
formulation of PSN, as they are not theoretically necessary, we use them as
additional inputs to the LSTM as they might help training. This construction is
illustrated in \cref{fig:construct}. New embeddings and distribution layers are
created upon encountering new addresses during training. In practice this is
implemented by sweeping through the samples used to calculate the gradient
estimator. It is similarly during these sweeps that new address transitions are
identified. For each address $a_{t}$ we construct $\mathcal{K}_{a_{t}}$ when new
address transitions are found. \cref{algo:psn-expand} is then used for each of
those addresses.

When replacing the reference simulator with the PSN, it is initialized using
$h_{0}$ and embeddings $x_0$, $d_0$, and $a_0$. These initial values are
typically set to zero, but could be learnable parameters. The unique first
address $a_1$ (which is guaranteed to be unique as the first point of
stochasticity in a program is always the same) is fed to the PSN and the
surrogate program starts its execution. At each subsequent time step $t$ the
PSN produces a sample $x_{a_t}$ and address $a_{t+1}$, which then propagates
the PSN forward where until an \texttt{end-execution} address is sampled. This
process is illustrated in~\cref{fig:construct}.

\section{Experiments}
\label{app:exp}

Here we provide various model, training, and validation specifications, along
with additional results and evidence that support the claims made in the main
paper.
\subsection{Model Specifications}

We largely use the default specifications found in PyProb~\citep{pyprob}. We
report the configurations whenever they differ from those default values.
We use the same configuration names found in PyProb, so that they can be
directly transferable from this paper. A description to each configuration will
be given the first time the configuration appears and only when the configuration
is not obvious (such as learning rate and optimizer).

\subsubsection{Stochastic Control Flow Experiment}
\label{app:model_spec_flow}

\cref{fig:toy_learning_curve} shows learning curves (training and validation)
for (a) the PSN and (b) the inference network. For this experiment we
continuously generate traces during training in an online
fashion. Therefore there is no risk of overfitting to a specific dataset and no
validation set is used.

\begin{figure*}[h!]
  \centering
  \includegraphics[scale=1.0]{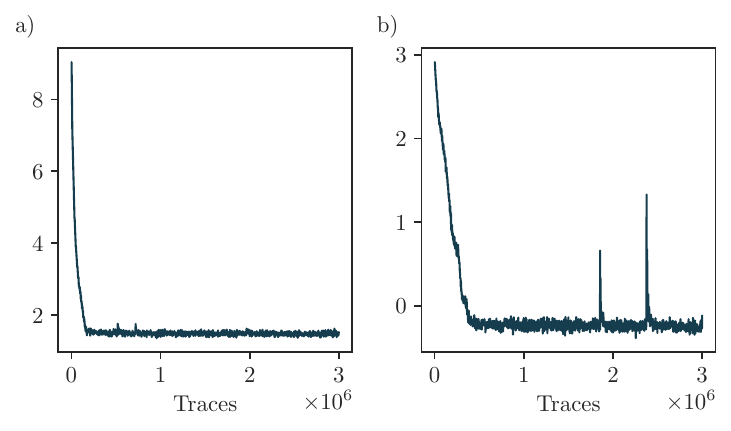}
  \caption{Learning curves for (a) the PSN and (b) the
    inference network associated with the stochastic control flow experiment.}
  \figlab{toy_learning_curve}
\end{figure*}

\begin{table*}[h!]
  \caption{Experiment configuration for the stochastic control flow experiment}
  \tablab{exp-config-toy}
  \centering
  \begin{tabular}{p{0.40\linewidth}p{0.14\linewidth}p{0.14\linewidth}p{0.2\linewidth}}
    \toprule
    Parameter/setting & IC & PSN & Description \\
    \midrule
    Optimizer & Adam & Adam & \\
    Learning rate & $\num{5d-4}$ & $\num{5d-4}$ & \\
    Training data size & 500,000 & 500,000 & \\
    Batch Size & 512 & 512 & \\
    \texttt{sample\_embedding\_dim} & 10 & 10 & The size of each variable embedding \\
    \texttt{address\_embedding\_dim} & 24 & 24 & The size of the address embedding which are learnable parameters \\
    \texttt{distribution\_type\_embedding\_dim} & 24 & 24 & The size of the distribution type embedding which are learnable parameters \\
    \texttt{observe\_embedding} & \{x: \{\text\{depth: 4, dim: 10, hidden\_dim: 10\}\}\} & N/A & \texttt{depth} is the number of linear layers mapping from the value $x$ each with \texttt{hidden\_dim} number of neurons. The output size (going into the LSTM) is \texttt{dim} \\
    \texttt{lstm\_depth} & 1 & 1 & Number of stacked LSTMs \\
    \texttt{lstm\_dim} & 150 & 150 & Size of hidden state in each LSTM \\
    \texttt{inf\_variable\_embedding} & \{theta: \{\text\{num\_layers: 2, hidden\_dim: 50\}\}\} & N/A & The names should be self-explanatory and are similar to \texttt{observe\_embedding} except the input to these layers are the output from the LSTM\\
    \texttt{surr\_variable\_embedding} & N/A & \{theta: \{\text\{num\_layers: 2, hidden\_dim: 50\}\}\} &  Same meaning as above but for the PSN \\
    \bottomrule
  \end{tabular}
\end{table*}

\clearpage

\subsubsection{Process Simulation of Composite Materials}
\label{app:model_spec_raven}
\cref{fig:raven_learning_curve} shows learning curves (training and validation)
for (a) the PSN and (b) the inference network. In this experiment we construct
a training set containing 200,000 traces which is iterated through until the number of traces
specified in~\tabref{exp-config-raven} has been encountered. The validation set
contains 7680 traces.

\begin{figure*}[h!]
  \centering
  \includegraphics[scale=1.0]{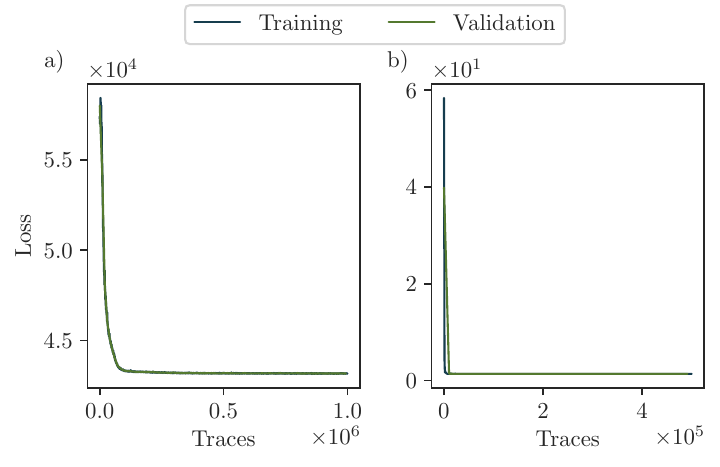}
  \caption{Training and validation learning curves for (a) the PSN and (b) the
    inference network associated with the process simulation of composite
    materials experiment.}
  \figlab{raven_learning_curve}
\end{figure*}

\begin{table*}[h!]
  \caption{Experiment configuration for the process simulation of composite materials experiment}
  \tablab{exp-config-raven}
  \centering
  \begin{tabular}{p{0.35\linewidth}p{0.28\linewidth}p{0.20\linewidth}}
    \toprule
    Parameter/setting & IC & PSN \\
    \midrule
    Optimizer & Adam & Adam \\
    Learning rate & $\num{d-3}$ & $\num{d-4}$ \\
    Training data size & 500,000 & 1,000,000 \\
    Batch Size & 256 & 256 \\
    \texttt{sample\_embedding\_dim} & 256 & 256 \\
    \texttt{address\_embedding\_dim} & 24 & 24 \\
    \texttt{distribution\_type\_embedding\_dim} & 24 & 24 \\
    \texttt{observe\_embedding} & \{temps\_bottom: \{depth: 2, dim: 500, hidden\_dim: 500\}, air\_temp\_bot: \{depth: 2, dim: 500, hidden\_dim: 500\}, air\_temp\_top: \{depth: 2, dim: 500, hidden\_dim: 500\}, temps\_config: \{dim: 10, hidden\_dim: 256\}\} & N/A \\
    \texttt{lstm\_depth} & 2 & 2 \\
    \texttt{lstm\_dim} & 512 & 512 \\
    \texttt{inf\_variable\_embedding} & \{config: \{\text\{hidden\_dim: 256\}\}\} & N/A  \\
    \texttt{surr\_variable\_embedding} & N/A & \{latent\_temps: \{\text\{num\_layers: 2, hidden\_dim: 500\}, temps\_config: \{hidden\_dim: 256\}\}\} \\
    \bottomrule
  \end{tabular}
\end{table*}

\clearpage

\subsubsection{Program synthesis Flow Experiment}
\label{app:model_spec_program}

\begin{figure*}[h!]
  \centering
  \includegraphics[scale=0.9]{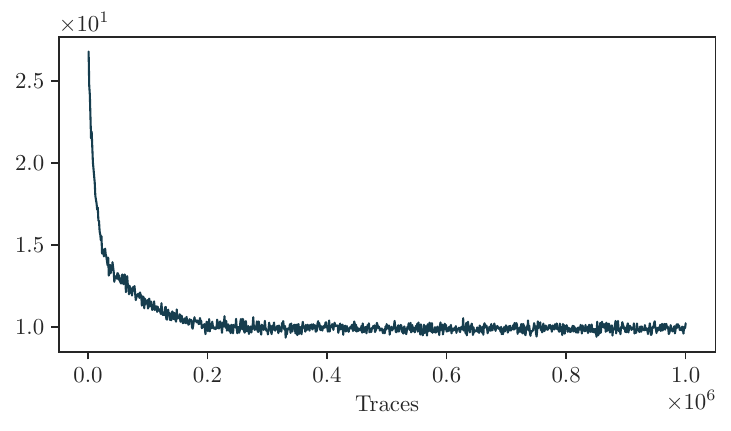}
  \caption{Learning curves for the PSN.}
  \label{synthesis_learning_curve}
\end{figure*}

The configurations used for training the surrogate in the program synthesis
experiment are the same as those found in \cref{tab:exp-config-toy}, while
\cref{synthesis_learning_curve} presents learning curves for the trained
surrogate.

\subsection{Running Times for Process Simulation of Composite Materials}
\label{app:running_times}

\begin{table}[h!]
  \caption{Runtime [traces/s] comparisons. We calculate the number of traces
    produced per second when (1) running just the simulator or PSN and (2) when
    performing SIS in either model. We see a slowdown in traces per second for
    the PSN when performing inference, as the inference engine adds additional
    overhead. However, as the simulator is
    considerably slower, it remains the computational bottleneck during
    inference. The reported run-times are achieved using an Intel(R) Xeon(R) CPU
    E3-1505M v5 @ 2.80GHz.}
  \tablab{runtime}
  \centering
  \begin{tabular}{llll}
    \toprule
    & Simulator ($t_{\mr{sim}}[\mr{traces}/\si{\second}]$)  & PSN ($t_{\mr{PSN}}[\mr{traces}/\si{\second}]$) & \textbf{Speedup} [$\nicefrac{t_{\mr{PSN}}}{t_{\mr{sim}}}$] \\
    \midrule
    PSN & 0.32 & 28.87 & \textbf{90.16}  \\
    IC in PSN & 0.31 & 4.75 & \textbf{15.32} \\
    \bottomrule
  \end{tabular}
\end{table}
\clearpage
\subsection{Results for the Process Simulation of Composite Materials Experiment}
\label{app:raven}

\begin{figure*}[h!]
  \centering
  \includegraphics[scale=0.81]{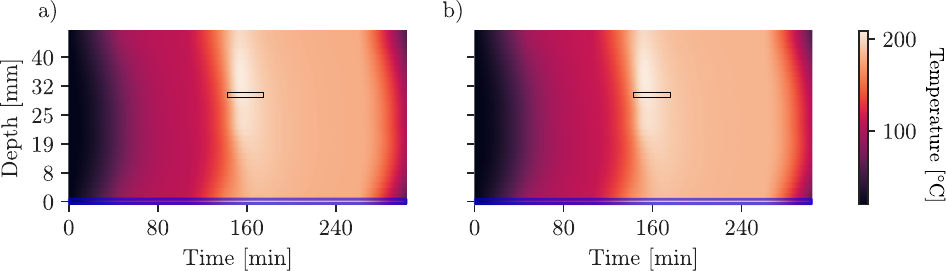}
  \caption{Illustration of a process simulation of composite materials. Each
    subfigure shows a temperature profile in degrees Celsius as a function of
    time along the $x$ axis and depth along the $y$-axis. (a) shows the output
    of the Convergent Composite material simulator
    RAVEN~\citep{Technologies2019}, simulating the curing process of a
    particular part. (b) shows the same process but originating from our
    \textit{probabilistic surrogate network}. We perform inference in this
    process, where we infer the expected temperature in a specific time window
    (black box) conditioned on observed surface temperature measurements (blue
    boxes).}
  \label{fig:guess}
\end{figure*}
\cref{fig:guess} compares output from our PSN and the reference simulator. As
these outputs are indistinguishable, it provides further evidence that our PSN
accurately models the reference simulator.

\clearpage
\subsection{Stochastic Control Flow Address Transitions}
\label{app:transitions}

\begin{figure}[h!]
  \centering
  \includegraphics[scale=1.0, trim={0 0 4.5cm 0}, clip]{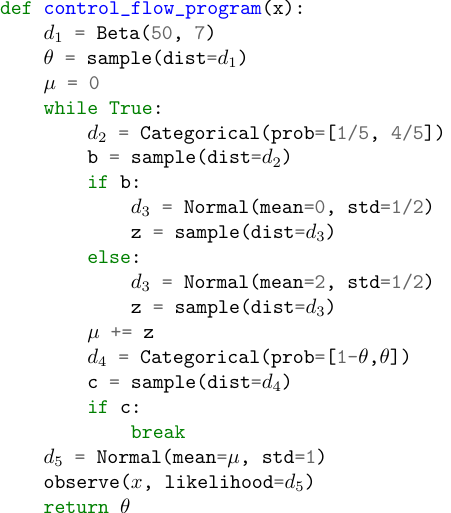}
  \caption{Program containing stochastic control flow in the form of a for-loop
    with a nested if-else statement. The task here would be to perform posterior
    inference of $\theta$ given the observed value of $x$.}
  \figlab{prog}
\end{figure}

For reference we re-illustrate the program~\cref{fig:prog} also shown in the
main paper. The program contains two nested layers of stochastic control flow,
allowing for an assessment of PSNs' capacity to learn the associated address
transitions.~\cref{fig:address-transitions}(a) and (b) complements the results
reported in the main paper by showing that the address transition paths and their associated estimated
probabilities (using 50,000 traces each) of the program and the trained PSN are near indistinguishable. Only for long traces does small deviations begin to appear. It is reasonable to expect slight discrepancies between
the address transition probabilities for increasingly long traces. The address
occurrence probability decreases exponentially in the number of times $n$ the
original program stays in the for-loop -- \ie $\theta^n$. Therefore we can
expect (with reasonable probability) either the PSN or the program to produce
addresses not produced by the other, when those addresses originate from
executions with large \texttt{for loop} iterations. We conclude that these
results show that the PSN indeed has learned accurate address transitions and
support the claim made in the main paper.

\begin{figure*}[h!]
  \centering
  \includegraphics[scale=0.95]{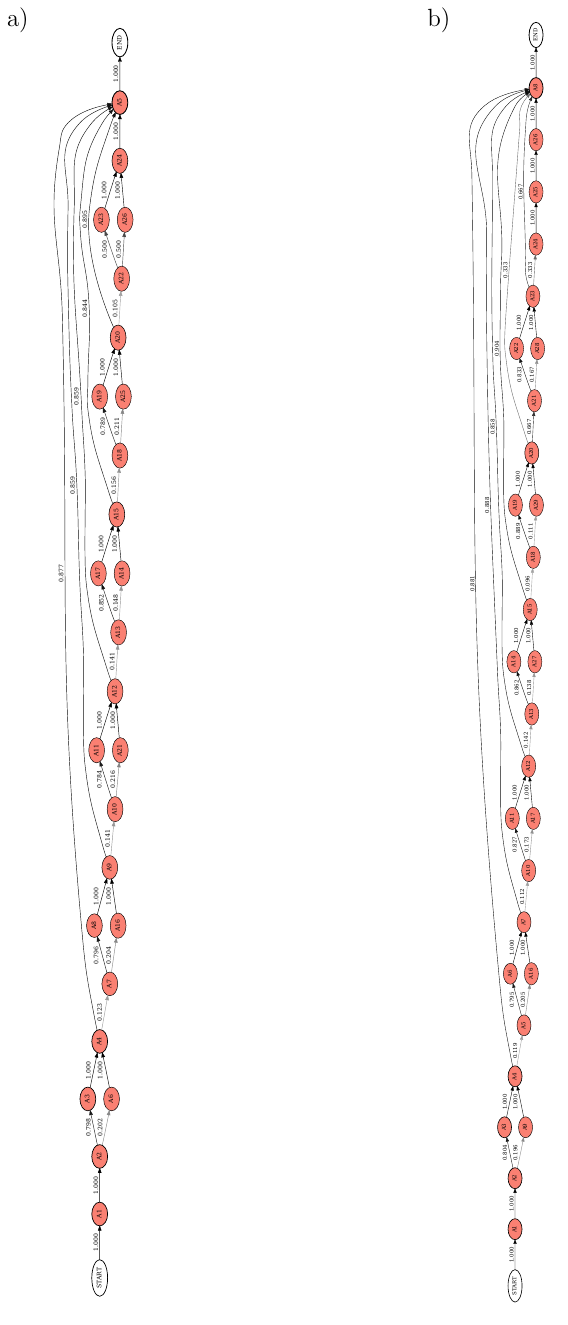}
  \caption{(a) Address transitions sampled from the original model shown
    in~\cref{fig:prog} with \tabref{orig_address_dict} mapping the address id
    A[i] to the actual address. (b) Address transitions sampled from the PSN,
    with \tabref{surr_address_dict} mapping the address id A[i] to the actual
    address allowing us to compare (a) and (b). For each plot the address
    transition probabilities are estimated across 50,000 traces.}
  \figlab{address-transitions}
\end{figure*}
\clearpage
\begin{table*}[h!]
  \caption{Address ID to address name for \cref{fig:address-transitions}.}
      \tablab{orig_address_dict}  \centering
  \begin{tabular}{ll}
    \toprule
    Address ID & Address \\
    \midrule
   A1 & \UScore{30__forward__theta__Beta__1}\\ 
   A2 & \UScore{bern_0__Categorical(len_probs:2)__1}\\ 
   A3 & \UScore{z_1_0__Normal__1}\\ 
   A4 & \UScore{c_0__Categorical(len_probs:2)__1}\\ 
   A5 & \UScore{280__forward__?__Normal__1}\\ 
   A6 & \UScore{z_2_0__Normal__1}\\ 
   A7 & \UScore{bern_1__Categorical(len_probs:2)__1}\\ 
   A8 & \UScore{z_1_1__Normal__1}\\ 
   A9 & \UScore{c_1__Categorical(len_probs:2)__1}\\ 
   A10 & \UScore{bern_2__Categorical(len_probs:2)__1}\\ 
   A11 & \UScore{z_1_2__Normal__1}\\ 
   A12 & \UScore{c_2__Categorical(len_probs:2)__1}\\ 
   A13 & \UScore{bern_3__Categorical(len_probs:2)__1}\\ 
   A14 & \UScore{z_2_3__Normal__1}\\ 
   A15 & \UScore{c_3__Categorical(len_probs:2)__1}\\ 
   A16 & \UScore{z_2_1__Normal__1}\\ 
   A17 & \UScore{z_1_3__Normal__1}\\ 
   A18 & \UScore{bern_4__Categorical(len_probs:2)__1}\\ 
   A19 & \UScore{z_1_4__Normal__1}\\ 
   A20 & \UScore{c_4__Categorical(len_probs:2)__1}\\ 
   A21 & \UScore{z_2_2__Normal__1}\\ 
   A22 & \UScore{bern_5__Categorical(len_probs:2)__1}\\ 
   A23 & \UScore{z_1_5__Normal__1}\\ 
   A24 & \UScore{c_5__Categorical(len_probs:2)__1}\\ 
   A25 & \UScore{z_2_4__Normal__1}\\ 
   A26 & \UScore{z_2_5__Normal__1}\\ 
   \bottomrule
  \end{tabular}
\end{table*}

\begin{table*}[h!]
  \caption{Address ID to address name for \cref{fig:address-transitions}.}
      \tablab{surr_address_dict}  \centering
  \begin{tabular}{ll}
    \toprule
    Address ID & Address \\
    \midrule
   A1 & \UScore{30__forward__theta__Beta__1}\\ 
   A2 & \UScore{bern_0__Categorical(len_probs:2)__1}\\ 
   A3 & \UScore{z_1_0__Normal__1}\\ 
   A4 & \UScore{c_0__Categorical(len_probs:2)__1}\\ 
   A5 & \UScore{bern_1__Categorical(len_probs:2)__1}\\ 
   A6 & \UScore{z_1_1__Normal__1}\\ 
   A7 & \UScore{c_1__Categorical(len_probs:2)__1}\\ 
   A8 & \UScore{280__forward__?__Normal__1}\\ 
   A9 & \UScore{z_2_0__Normal__1}\\ 
   A10 & \UScore{bern_2__Categorical(len_probs:2)__1}\\ 
   A11 & \UScore{z_1_2__Normal__1}\\ 
   A12 & \UScore{c_2__Categorical(len_probs:2)__1}\\ 
   A13 & \UScore{bern_3__Categorical(len_probs:2)__1}\\ 
   A14 & \UScore{z_1_3__Normal__1}\\ 
   A15 & \UScore{c_3__Categorical(len_probs:2)__1}\\ 
   A16 & \UScore{z_2_1__Normal__1}\\ 
   A17 & \UScore{z_2_2__Normal__1}\\ 
   A18 & \UScore{bern_4__Categorical(len_probs:2)__1}\\ 
   A19 & \UScore{z_1_4__Normal__1}\\ 
   A20 & \UScore{c_4__Categorical(len_probs:2)__1}\\ 
   A21 & \UScore{bern_5__Categorical(len_probs:2)__1}\\ 
   A22 & \UScore{z_1_5__Normal__1}\\ 
   A23 & \UScore{c_5__Categorical(len_probs:2)__1}\\ 
   A24 & \UScore{bern_6__Categorical(len_probs:2)__1}\\ 
   A25 & \UScore{z_1_6__Normal__1}\\ 
   A26 & \UScore{c_6__Categorical(len_probs:2)__1}\\ 
   A27 & \UScore{z_2_3__Normal__1}\\ 
   A28 & \UScore{z_2_5__Normal__1}\\ 
   A29 & \UScore{z_2_4__Normal__1}\\ 
   \bottomrule
  \end{tabular}
\end{table*}

\clearpage

\subsection{Program Synthesis Details}
\label{app:prog-synthesis-details}
\begin{figure*}[h!]
  \centering
  \includegraphics[scale=0.55]{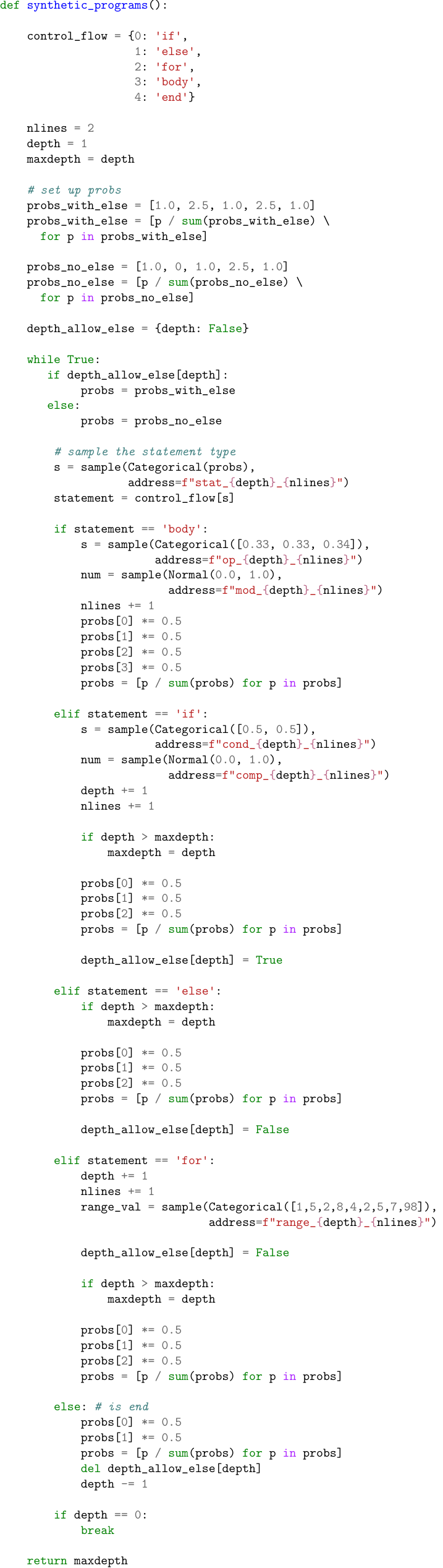}
  \caption{Model describing the program synthesis generative model.}
  \label{fig:simulator}
\end{figure*}

\begin{figure*}[h!]
  \centering
  \includegraphics[scale=1.0]{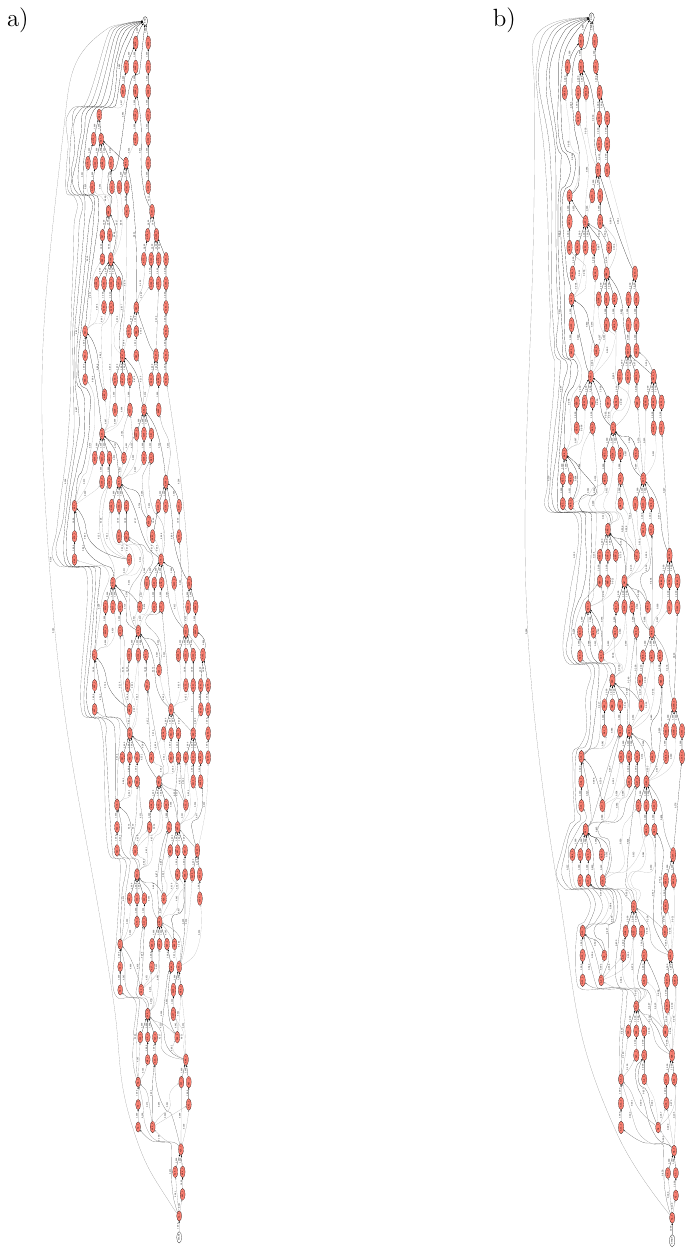}
  \caption{(a) Address transitions sampled from the original model shown
    in~\cref{fig:simulator} (b) Address transitions sampled from the PSN. For
    each plot the address transition probabilities are estimated across 50,000
    traces.}
  \label{fig:address_transitions}
\end{figure*}

\begin{figure*}[h!]
  \centering
  \includegraphics[scale=0.9]{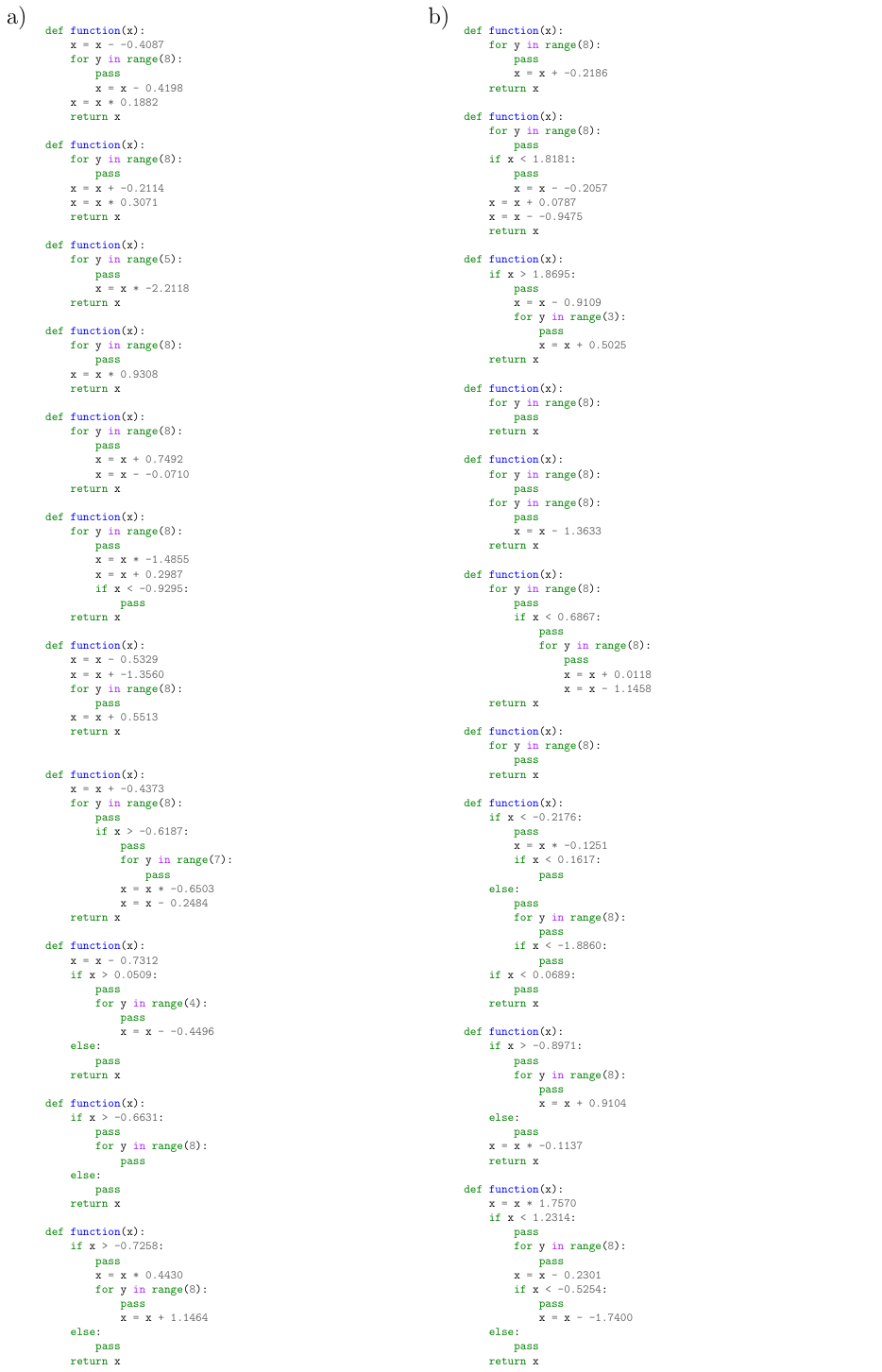}
  \caption{(a) Example programs sampled from the original model shown
    in~\cref{fig:simulator} (b) Example programs sampled from the learned
    surrogate.}
  \label{fig:example_programs}
\end{figure*}

The python code describing the generative model we approximate with a surrogate
is given in \cref{fig:simulator}. Note that the $\texttt{depth\_allow\_else}$
data structure is in effect a stack that keeps track of the nesting of
$\texttt{if}$ and $\texttt{else}$ statements. To generate valid programs, the
surrogate has to learn that valid programs can only sample an $\texttt{else}$
statement if an $\texttt{if}$ statement has preceded it on the same nesting
level. Furthermore, in our generative model, a valid program can only end at the
lowest nesting level. Expanding on the results presented in the main text,
additional example programs for both the original and the surrogate are
displayed in \cref{fig:example_programs}. Address transitions for the synthetic
programs can be found in \cref{fig:address_transitions}. The structure of these
transitions makes it clear that the program can only finish from specific
addresses, corresponding to those sampled at the lowest nesting level. It is
evident from the transitions presented for the surrogate that these dependencies
are accurately captured.

\end{document}